\documentclass[review,11pt]{ReportTemplate}
\usepackage{bm}
\usepackage{graphicx}
\usepackage{paralist,algorithmic,algorithm}
\usepackage{amsmath,mathrsfs,amssymb}
\usepackage{ textcomp }
 \usepackage[american]{babel}
\usepackage{algorithm}
\usepackage{algorithmic}
  \usepackage{subfigure}
   \usepackage{epsfig}
\newtheorem{defn}{Definition}[section]
 \DeclareMathOperator*{\argmin}{arg\,min}
 \DeclareMathOperator*{\argmax}{arg\,max}

\def \x {\mathbf{x}}

\begin{document}
\begin{frontmatter}
\title{Classification under\\ Streaming Emerging New Classes:\\ A  Solution using Completely Random Trees}

\author{Xin Mu$^{\star}$}
\author{Kai Ming Ting$^{\sharp}$}
\author{Zhi-Hua Zhou$^{\star}$\corref{cor1}}
\address{$^{\star}$National Key Laboratory for Novel Software Technology\\
Nanjing University, Nanjing 210093, China} \cortext[cor1]{\small Corresponding author.
Email: zhouzh@nju.edu.cn}
\address{$^{\sharp}$ School of Engineering and Information Technology, \\
Federation University, Victoria, Australia.}

\begin{abstract}
This paper investigates an important problem in stream mining, i.e., classification under streaming emerging new classes or \emph{SENC}. The common approach is to treat it as a classification problem and solve it using either a supervised learner or a semi-supervised learner. We propose an alternative approach by using unsupervised learning as the basis to solve this problem. The \emph{SENC} problem can be decomposed into three sub problems: detecting emerging new classes, classifying for known classes, and updating models to enable classification of instances of the new class and detection of more emerging new classes.
The proposed method employs completely random trees which have been shown to work well in unsupervised learning and supervised learning independently in the literature. This is the first time, as far as we know, that completely random trees are used as a single common core to solve all three sub problems: unsupervised learning, supervised learning and model update in data streams. We show that the proposed unsupervised-learning-focused method often achieves significantly better outcomes than existing classification-focused methods.
\end{abstract}

\end{frontmatter}

{T}{his} paper investigates an important problem in data streams, i.e., classification under streaming emerging new class or \emph{SENC}. In many real-world data mining problems, the environment is open and changes gradually. In the streaming classification problem, some new classes are likely to emerge as the environment changes.
The predictive accuracy of a previously trained classifier will be severely degraded if it is used to classify instances of a previously unseen class in the data stream.
Ideally, we would like instances of a new class to be detected as soon as they emerge in the data stream; and only instances which are likely to belong to known classes are passed to the classifier to predict their classes.

It is assumed that true class labels are not available throughout the entire process, except a training set of known classes which is used to train a classifier (and a detector for new classes) at the beginning of the data stream. After the deployment of the classifier (and the detector), any future updates of the models must rely on the unlabelled instances as they appear in the data stream.
Note that this assumption does not prevent the proposed method from using true class labels when they are available. It sets the hardest condition in the \emph{SENC} problem.

\begin{figure}
   \centering
 \epsfig{file=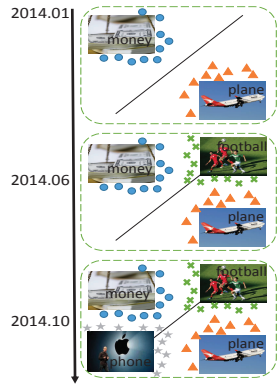,height=2in,width=2in}
  \caption{Image classification in a data stream}
 \label{fig_example}
 \vspace{-5mm}
\end{figure}

An illustrative example is provided in Figure \ref{fig_example} which shows a news image classifier system making predictions in a data stream. Assume that a classifier about news content is built in early 2014, which starts with two classes (money and airplane); then some new classes (football and phone) emerge in two later periods in the data stream. The system must have the ability to detect those new classes and update itself timely in order to maintain the predictive accuracy.  

Conceptually, the \emph{SENC} problem can be decomposed into three sub problems: detecting emerging new classes, classifying known classes, and updating models to enable classification of instances of the new classes and detection of more emerging new classes. For every test instance in a data stream, the detector acts as a filter to determine whether it is likely to belong to a known class. If it is, the instance is passed on to the classifier to produce a class prediction. Otherwise, the instance is declared a new class and placed in a buffer which stores candidates of previously unseen class. When the candidates have reached the buffer size, they are used to update both the classifier and the detector. The process repeats in the data stream after the models are updated.

The overall aim of the task is to maintain high classification accuracy continuously in a data stream. Thus, the challenges in the \emph{SENC} problem are to detect emerging new classes and classify instances of known classes with high accuracy, and to perform model update efficiently in data streams. In order to maintain the model complexity to a reasonable size, model components related to currently inactive classes must be eliminated from the current model.

We show that these challenges can be met by using completely random trees, and the proposed method often achieves significantly better outcomes than existing more complicated methods. The proposed method has the following distinguishing features:
\begin{itemize}
\item The proposed method employs an unsupervised learning method as the basis to solve the \emph{SENC} problem, and has a single common core which acts as distinct unsupervised learner and supervised learner. In contrast, most existing methods treat this problem as a classification problem and employ a supervised or semi-supervised learning approach~\cite{ma2003time, DBLP:conf/aaai/DaYZ14} to solve it.
\item The method explicitly differentiates anomalies of known classes from instances of emerging new classes using an unsupervised learning anomaly detection approach.
\item The model is updated without the initial training set because the proposed method does not need to train new models for every future model updated. In contrast, most existing methods must keep this training set in order to train new models (e.g., LACU-SVM~\cite{DBLP:conf/aaai/DaYZ14}.)
\end{itemize}

Note that most of the existing methods mentioned above are designed to solve part of the \emph{SENC} problem only. Details are provided in Section \ref{sec_related_work}.

Our main contribution is the proposal to shift the focus of treating \emph{SENC} as a classification problem to one based on unsupervised anomaly detection problem. In other words, the focus is shifted from the second sub problem to the first sub problem which is more critical in solving the entire problem. This shift brings about an integrated approach to solve all three sub problems in \emph{SENC}. No such solution exists in the current classification-focused approaches, as far as we know.

The rest of this paper is organized as follows: Section 1 describes the intuition of the proposed algorithm. Section \ref{sec_related_work} reviews the related work. Section 4 and 5 describe related definitions and  the details of the proposed algorithm. We report the experimental results in Section 6. The conclusion is provided in the last section.

\section{The intuition}

\subsection{Detecting emerging new classes}

The intuition is that anomalies of known classes are at the fringes of the data cloud of known classes, and instances of any emerging new classes are far from the known classes.
To detect emerging new classes, we propose to treat instances of any new class as ``outlying'' anomalies which are significantly different from both instances and anomalies of the known classes.

The anomaly detector for the \emph{SENC} problem must be able to differentiate between these two types of anomalies. The assumption is that anomalies of the known classes are more ``normal'' than the ``outlying'' anomalies. This is a reasonable assumption in this context because only instances of the known classes are available to train the anomaly detector.

An anomaly detector often categorises the feature space into two types of regions: anomaly and normal. Following the above idea, we propose to further subdivide each anomaly region into two sub regions: ``outlying'' anomaly sub region and anomaly sub region: (1) The instances in \emph{anomaly sub region} is closer to the region of normal instances than instances from emerging new classes as the anomalies and normal instances are generated from the same distribution. (2) \emph{``Outlying'' anomaly sub region} is further away from the normal region and anomaly sub region. A test instance is regarded as belonging to an emerging new class if it falls in the ``outlying''  anomaly sub region.

\begin{figure}
\centering
  {
   \epsfig{file=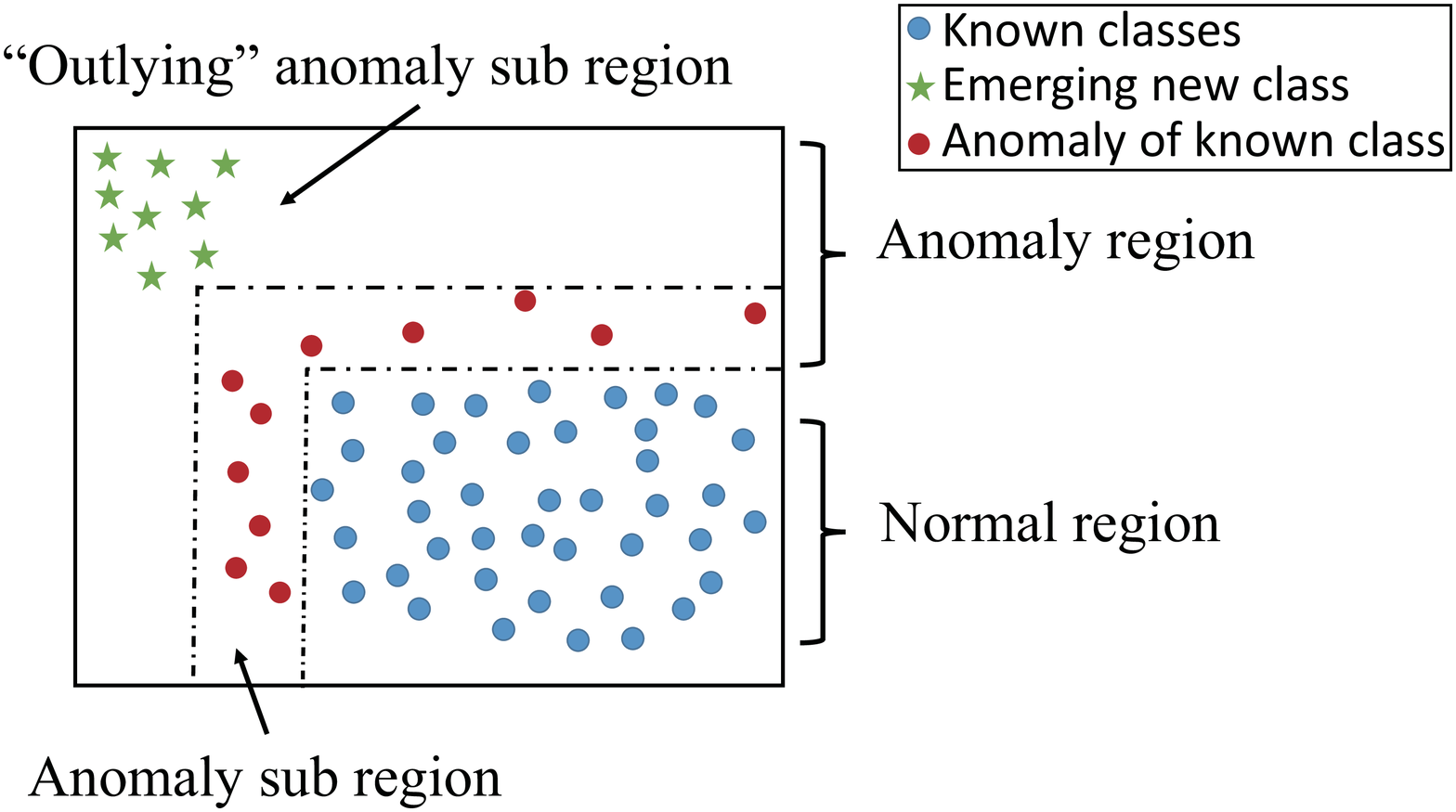,height=1.5in,width=2.8in}}
  \caption{An illustration to build an ``outlying'' anomaly sub region} 
 \label{fig_subregion}
  \vspace{-3mm}
\end{figure}

Figure \ref{fig_subregion} illustrates the normal and anomaly regions constructed by an anomaly detector. The anomaly region is further partitioned into two sub regions. The sub region outside the anomaly sub region is the ``outlying''  anomaly sub region.

The construction of ``outlying'' anomaly sub regions assumes that anomaly regions can be identified. We show in Section \ref{sec_threhold} that this can be easily achieved using a threshold of the anomaly scores provided by an anomaly detector to categorise all regions into two types: anomaly and normal.

\subsection{Classification and efficient model update}

If we treat the second sub problem, i.e., classification, as having no relation to the first sub problem for detecting emerging new classes, then any classifier can be applied. However, in order to facilitate efficient model update that enables classification of newly detected class and detection of more emerging new classes in data streams, we suggest an integrated approach which has a single common core for both the detection and classification tasks.

An unsupervised learner iForest\cite{liu2008isolation}, which induces completely random trees, has enabled us to implement the integrated approach with ease. This is because previous works \cite{Israndommodelbetter, MaximizingTree} have shown that, ensemble of completely random trees \cite[Chap.3.5]{zhou2012ensemble}, as an extreme case of variable-random trees \cite{liu2008spectrum}, can be successfully applied as a powerful classifier. 
We use exactly the same completely random trees, generated for the purpose of anomaly detection, for classification. This can be easily achieved by simply recording the class labels (provided in the training set) in each leaf. This is the only additional step that needs to be done in the training process to produce an ensemble of completely random trees that will act as both an unsupervised learner (to detect emerging new classes) and an supervised learner (to classify known classes) in data streams.

As the single core for both tasks is completely random trees only, they can be updated easily when a sufficient number of instances of emerging new classes have been detected. The single core also facilitates to maintain the model complexity in a reasonable size by using effective model retiring mechanism and growing mechanism in the model update process.

In a nutshell, we introduce a simple and unique method to solve the \emph{SENC} problem and show that the proposed method can detect emerging new classes and classify known classes with high accuracy, and perform model update efficiently in data streams. Our empirical evaluation shows that it often performs significantly better than existing more complex methods.


 \section{Related work}
\label{sec_related_work}

The \emph{SENC} problem has the following challenges:
\begin{enumerate}
\item In the extreme case, no true labels except in the initial training set, i.e, true labels are not available after the model deployment.
\item A prediction must be made immediately for each incoming instance in the stream. 
\item Store no data permanently from the data stream.
\item Fast model update.
\end{enumerate}
Note that, as far as we know, there is no an algorithm that using one single core to conquer the whole \emph{SENC} challenges. We review the related work with respect to these challenges as following.

Class-incremental learning (C-IL) \cite{zhou2002hybrid} is a branch of incremental learning which modifies a previously trained classifier to deal with emerging new classes. It has been found to be useful in various applications, e.g., detecting bots \cite{DBLP:conf/kdd/ChenRT11}, face recognition \cite{DBLP:conf/iccv/HuangAYLK07} and video concept detection\cite{yang2007cross}. C-IL problems includes open set recognition\cite{scheirer2013toward}, Learning with Augmented Class (LAC)\cite{DBLP:conf/aaai/DaYZ14}. All of these works are in the batch mode setting.
The \emph{SENC} problem is a C-IL problem in the data stream context.

In addition, many existing methods treat the \emph{SENC} problem as a classification problem.
This is the reason why they have employed supervised learning or semi-supervised learning approaches. Moreover, most of these studies assume that instances of an emerging new class are identified by some other mechanism and focuses on methods to train and incorporate classifiers which can classify new classes incrementally with previously trained classifiers\cite{DBLP:conf/aaai/DaYZ14,  kuzborskij2013n}. As a result, no existing methods in C-IL meet the four challenges mentioned above.

Learning with Augmented Class (LAC) \cite{DBLP:conf/aaai/DaYZ14} is a new effort for C-IL and addresses a research gap, i.e., to produce a detector for emerging new classes. Utilising unlabelled instances through semi-supervised learning, LACU-SVM \cite{DBLP:conf/aaai/DaYZ14} modifies a previously trained classifier to identify emerging new classes. Assuming the set of unlabelled instances containing sufficient instances of an emerging new class, a trained LACU-SVM can then assign a test instance to either one of the known classes or emerging new class. While it solves the first and second sub problems,
 it is a batch-mode method that requires to store all training data. Thus, it is not suitable in data streams and does not meet the four challenges.

The aim of \emph{novel class detection} is to identify new data which are not previously seen by a machine learning system during training. This is the first sub problem of \emph{SENC}. An example of this work in Bioinformatics \cite{DBLP:conf/wob/SpinosaC04} employs an one-class SVM approach to detect novel classes.
It is interesting to note that this approach does not make a distinction between novel class detection and \emph{anomaly detection} (or outlier detection) \cite{chandola2009anomaly}, which is the identification of items, events or observations which do not conform to an expected pattern in a data set in batch mode. It thus also does not meet the four \emph{SENC} challenges in data streams.

The goal of \emph{change point detection} is to detect changes in the generating distributions of the time-series. Many works have been conducted to tackle this problem \cite{DBLP:journals/automatica/Poor96} which include parametric methods \cite{desobry2005online} and non-parametric methods \cite{brodsky1993nonparametric}.
This problem is equivalent to the first sub problem in \emph{SENC}, without addressing the classification and model update issues.
Yet, others have focused on classification in data streams \cite{DBLP:conf/kdd/BifetHPKG09, Jin:2003:EDT:956750.956821, kolter2007dynamic}, without addressing the emerging new classes problem.

Another related work, ECSMiner, \cite{a5453372} tackles the novel class detection and classification problems by introducing time constraints for delayed classification. ECSMiner assumes that true labels of new emerging class can be obtained after some time delay; otherwise, models cannot be updated. In contrast, our proposed method assumes that no labels are available for the entire duration of a data stream.

The \emph{SENC} problem can be solved by treating the first two sub problems independently by using existing methods, i.e., a new class detector and a known classes classifier. To detect emerging new class, existing anomaly detectors (such as LOF \cite{DBLP:conf/sigmod/BreunigKNS00},  iForest \cite{liu2008isolation} and one-class SVM \cite{ma2003time}) can be employed; and multi-class SVM \cite{CC01a}) can be used as an the classifier for known classes.  In addition, existing supervised or semi-supervised batch  classification methods can be
 adapted to solve the \emph{SENC} problem, e.g., One-vs-rest SVM \cite{Rifkin:2004:DOC:1005332.1005336} and LACU-SVM \cite{DBLP:conf/aaai/DaYZ14}.

 However, all these algorithms do not solve the \emph{SENC} problem satisfactorily. Table \ref{tbl_data1} summarizes the ability of these algorithms and the proposed \emph{SENCForest} to meet the four challenges.

\begin{table}[h]
  \centering
  \caption{Ability of algorithms to meet the challenges of the \emph{SENC} problem.}
  \label{tbl_data1}
  \begin{tabular}{|c|c|c|c|c|}
    \hline
     Algorithm       & \multicolumn{4}{c|}{Challenge} \\ \cline{2-5}
            &  1   &  2  & 3 & 4   \\
       \hline
      LOF+SVM            &  $\times$  &  $\checkmark$  & $\times$ & $\times$ \\
        \hline
      1SVM+SVM  & $\times$   & $\checkmark$  & $\times$ & $\times$ \\
        \hline
     One-vs-rest SVM &  $\times$  & $\checkmark$  & $\times$ & $\times$ \\
        \hline
     LACU-SVM        &  $\times$  &  $\checkmark$ & $\times$ & $\times$ \\
    \hline
     iForest+SVM     &  $\times$  &  $\checkmark$ & $\times$ & $\checkmark$ \\
        \hline
     ECSMiner          &  $\times$  &  $\times$ & $\checkmark$ & $\checkmark$ \\
        \hline
     \emph{SENCForest}         &  $\checkmark$  &  $\checkmark$ & $\checkmark$ & $\checkmark$ \\
        \hline
  \end{tabular}
\end{table}

Details about those algorithms implemented and the proposed \emph{SENCForest} are provided in following sections.


\emph{SENCForest} is the only one which can meet all four challenges. Only ECSMiner, among existing algorithms, can meet Challenge \#3. Note that all existing algorithms assume that true labels are made available after the model deployment at some points in time---unable to meet Challenge \#1.

\section{Terminology Definition}
Before introducing the detail of our proposed algorithm, we will give the formal definitions of many
important concepts used in this paper.

\begin{defn}{Classification  under  Streaming Emerging new Class (SENC) problem:}
Given a training data set $D = \{(x_i,y_i)\}^L_{i=1}$, where $x_i \in R^d$ is a training instance and $y_i \in Y =\{1, 2,\ldots,K\}$ is the associated class label. A streaming data $S= \{(x'_t,y'_t)\}^\infty_{t=1}$, where $x' \in R^d$, $y' \in Y'=\{1, 2,\ldots, K, K +1,\ldots, M\}$ with $M> K$. The goal of learning with the SENC problem is to learn a model $f$ with $D$ initially; then $f$ is used as a detector for emerging new class and a classifier for known class. $f$  is updated timely such that it maintains accurate predictions for known and emerging new classes on streaming data $S$.
   \label{define:AK}
 \end{defn}

The \emph{SENC} problem can have different variations.  The hardest condition is when true class  labels are not available throughout the entire process, except that the initial training set of known classes is used to train a classifier (and a detector for new classes) at the beginning of the data stream. A relaxation of this condition produces easier \emph{SENC} problems. For example, true class labels are available at some intervals in streaming data $S$. In this paper, we show that the proposed method can deal with the hardest condition (in Section \ref{sec_results}) as well as some easier conditions (in Section \ref{sec_results_long}).


\begin{defn}{Scores for test instances:}
Model $f$ yields a score for a test instance $x$, which determines $x$ as belonging to either a known class or an emerging new class (i.e., an ``outlying'' anomaly.)
\label{define:New Score}
 \end{defn}

 \begin{defn}{Known Class Region and Anomaly Region:}
 Based on the score from $f$, the feature space is divided into two types of regions : (a) known class regions $K$ which have score $\ge \hat{\tau}$, (b) anomaly regions $A$ which have score $< \hat{\tau}$, where $\hat{\tau}$ is a threshold.
 \label{define:AK}
 \end{defn}


\begin{defn}{Anomalies of Known Classes:}
Let ${\cal O} = \{x_1,\dots,x_n\}$ be the training instances in an anomaly region $A$. The center of ${\cal O}$ is defined as $c = \frac{1}{n} \sum_{x \in {\cal O}} x$. Let $e \in {\cal O}$ be the farthest instance from $c$.  A ball $B$ centered at $c$ with radius $r = dist(c,e)$ is an anomaly sub region. Instances which fall into anomaly sub regions are Anomalies of Known Classes.
\label{define:Anomaly of Known Class}
 \end{defn}

\begin{defn}{Instances of an emerging new class} are ``outlying'' anomalies: $Q = A \backslash B$.
\label{define:Outlying Anomalies}
 \end{defn}

\section{The Proposed Algorithm}
In this section, we propose an efficient algorithm to deal with the \emph{SENC} problem named \emph{SENCForest} which is composed of \emph{SENCTrees} and assigns each instance, as it appears in a data stream, a class label: Emerging New Class or one of the known classes. Instead of treating it as a classification problem, we formulate it as a new class detection problem and solve it using an unsupervised anomaly detector as the basis to build \emph{SENCForest} which will finally act as both unsupervised learner and supervised learner.

We provide an overview of the procedure in section \ref{overview}. The pertinent details in the procedure are then provided in the following three sections.

\subsection{SENCForest: An Overview}
\label{overview}
\emph{SENCForest} has four major steps:

{\bf 1. Train a detector for emerging new classes}. Given the initial training set of known classes $D$, an unsupervised anomaly detector \emph{SENCForest} is trained, ignoring the class information, as follows:
\begin{enumerate}
\item Build an iForest \cite{liu2008isolation}.
\item Determine the path length\cite{liu2008isolation} threshold $\hat{\tau}$.\label{item_steps}
\item Within each region $A$, construct ball $B$ which covers all training instances which fall into this region. The area of the ball $B$ is anomaly sub region $A$. Any test instances which fall into $B$ are regarded as anomalies of known classes; those that fall outside $B$ are regarded as instances from an emerging new class.
\end{enumerate}

\begin{figure}
  \centering
  {
   \epsfig{file=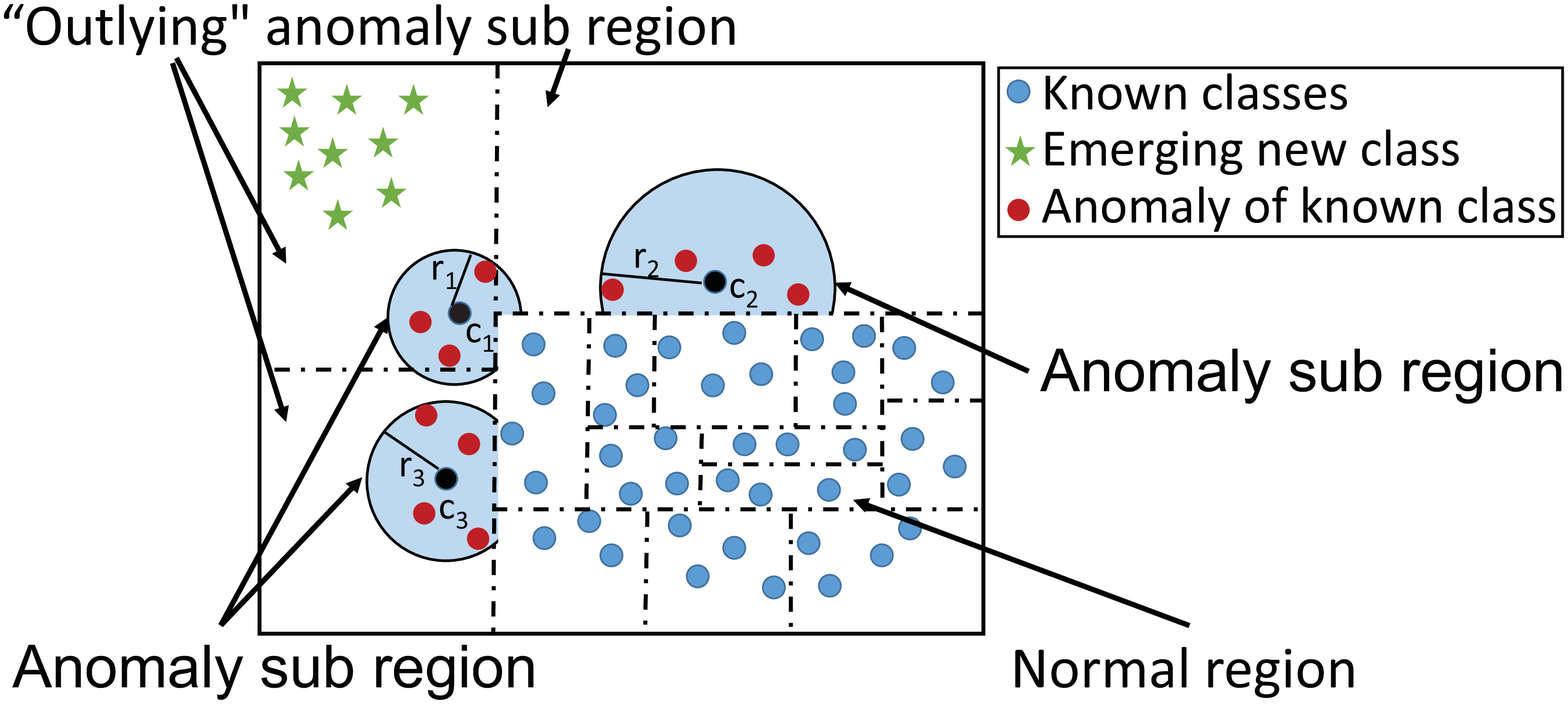,height=1.5in,width=17in}}
  \caption{An illustration to build an ``outlying'' anomaly sub region} 
 \label{fig_ifo}
  \vspace{-3mm}
\end{figure}

 The path length is introduced in iForest\cite{liu2008isolation}, which can be regard as an anomaly score for determining known class region and anomaly region(like Definition \ref{define:AK}). After training the detector of \emph{SENCForest}, model \emph{SENCForest} can yields a new class score  for a test instance $x$ through aggregating results of each tree in \emph{SENCForest}. Detail of iForest will be described in the following section.

 Figure \ref{fig_ifo} illustrates the regions constructed by an iTree which has axis-parallel boundaries, and the additional subdivision employs a ball to partition each anomaly region
into two sub regions. The anomaly sub region outside the ball is the ``outlying'' anomaly sub region.

{\bf 2. Using known class information to build a classifier from a detector}. Once the above new class detector is constructed, class distributions based on known class labels are recorded in each $K$ or $B$ region. Each region with class distribution acts as a classifier that outputs the majority class as the classification result for a test instance which fall into the region.

The training set is discarded once the training process is completed.

{\bf 3. Deployment in data stream}. \emph{SENCForest} is now ready to be deployed in a data stream, and it is assumed that no true class labels are available for model updated throughout the entire data stream. An instance in the data stream is given a class prediction by \emph{SENCForest} if it falls into $K$ or $B$ region; otherwise, it is identified as an instance from an emerging new class and placed in a buffer of size $s$.

{\bf 4. Model update}. The model update process in \emph{SENCForest} is simple. It begins when the buffer is full. Using instances from the buffer, the same tree growing process is then applied to each leaf of every existing tree until the stopping criterion is satisfied. The rest of the model update process follows the same steps from 1.\ref{item_steps} onwards, as described above. Note that the update largely involves newly grown subtrees, i.e., replacing leaf nodes which have the number of instances more than a set limit after taking new instances from the buffer into consideration. Thus, the whole process can be completed quickly. To maintain model size, mechanisms to retire 
\emph{SENCForest} are also employed in the model update process.

Section \ref{sec_enc} describes the pertinent details of training \emph{SENCForest} as both unsupervised detector and supervised learner. Deploying \emph{SENCForest} and model update in data streams are provided in Section \ref{sec_deployment}  and Section \ref{sec_Update}, respectively.

\subsection{SENCForest: Training process}
\label{sec_enc}

The training procedure to build an \emph{SENCForest} with both detection and classification functions  is detailed in Algorithms \ref{top level} and \ref{ACLTree}. These are the combined step to build iForest\cite{liu2008isolation} and to produce a classifier from a detector. The trees are then used to determine the path length threshold and to construct ``outlying'' anomaly regions described following respectively.
Note that the procedure is the same as in building iForest, except in line 2 of Algorithm \ref{ACLTree}. As the trees constructed are not exactly iTrees, we name the trees with the new classification capability, \emph{SENCTrees}.

{\bf Build an iForest}. The unsupervised anomaly detector \emph{iForest} \cite{liu2008isolation} is an ensemble of \emph{Isolation Tree (iTrees)}. ``Isolation'' is a unique concept in anomaly detection, as each iTree is built to isolate every instance from the rest of the instances in the training set. The idea is based on the fact that since anomalies are `few' and `different', they are more susceptible to isolation than normal instances. Hence, an anomaly can be isolated using fewer partitions in an iTree than a normal instance.

Liu et. al. \cite{liu2008isolation} show that iTrees can be created using a completely random process to achieve the required isolation. Given a random subsample of size $\psi$, a partition is produced by randomly selecting an attribute and its cut-point between the minimum and maximum values in the subsample. To produce an iTree, the partitioning process is repeated recursively until every instance in the subsample is isolated. An iForest is an ensemble of $z$ iTrees, each generated using a subsample randomly selected from the given training set.

In the testing process, an instance having a short path length, which is the number of edges it traversed from the root node to a leaf node of an iTree, is more like to be an anomaly. The average path length from all iTrees is used as the anomaly score for each test instance.

For both instances of emerging new class and anomalies of known classes, iForest will produce short path lengths because they all are individually `few' and `different' from the known classes. In order words, they are all in the regions with short path length in iTrees. We called this type of region, anomaly region $A$ to differentiate them from normal region $K$ which have long path length.

In order to detect emerging new class,  we first need to determine a path length threshold to differentiate $A$ from $K$. Then, build a sub region $B$ in each $A$ region which covers all training instances in the region. As these instances are from known classes, they are anomalies of known classes. These two processes are described in the following paragraph.

{\bf Determine the path length threshold.}
\label{sec_threhold}
As each region in iTree has its own path length, and anomaly regions $A$ are expected to have shorter path length than that from normal regions $K$, we employ the following method to determine the path length threshold to separate these two types of regions.

We produce a list $L$ which orders all path lengths representing all regions in an iTree in ascending order. A threshold $\tau$ in this list yields two sub-lists $L^l$ and $L^r$.
To find the best threshold, we use the following criterion which minimises the difference in standard deviations $\sigma(.)$:
\[ \hat{\tau} =  \argmin_\tau \  |\sigma(L^{r}) - \sigma(L^{l})| \]
The threshold $\hat{\tau}$ is used to differentiate anomaly regions $A$ from normal regions $K$, where the former has low path length and the latter has long path length.

\begin{figure}
  \centering
  \epsfig{file=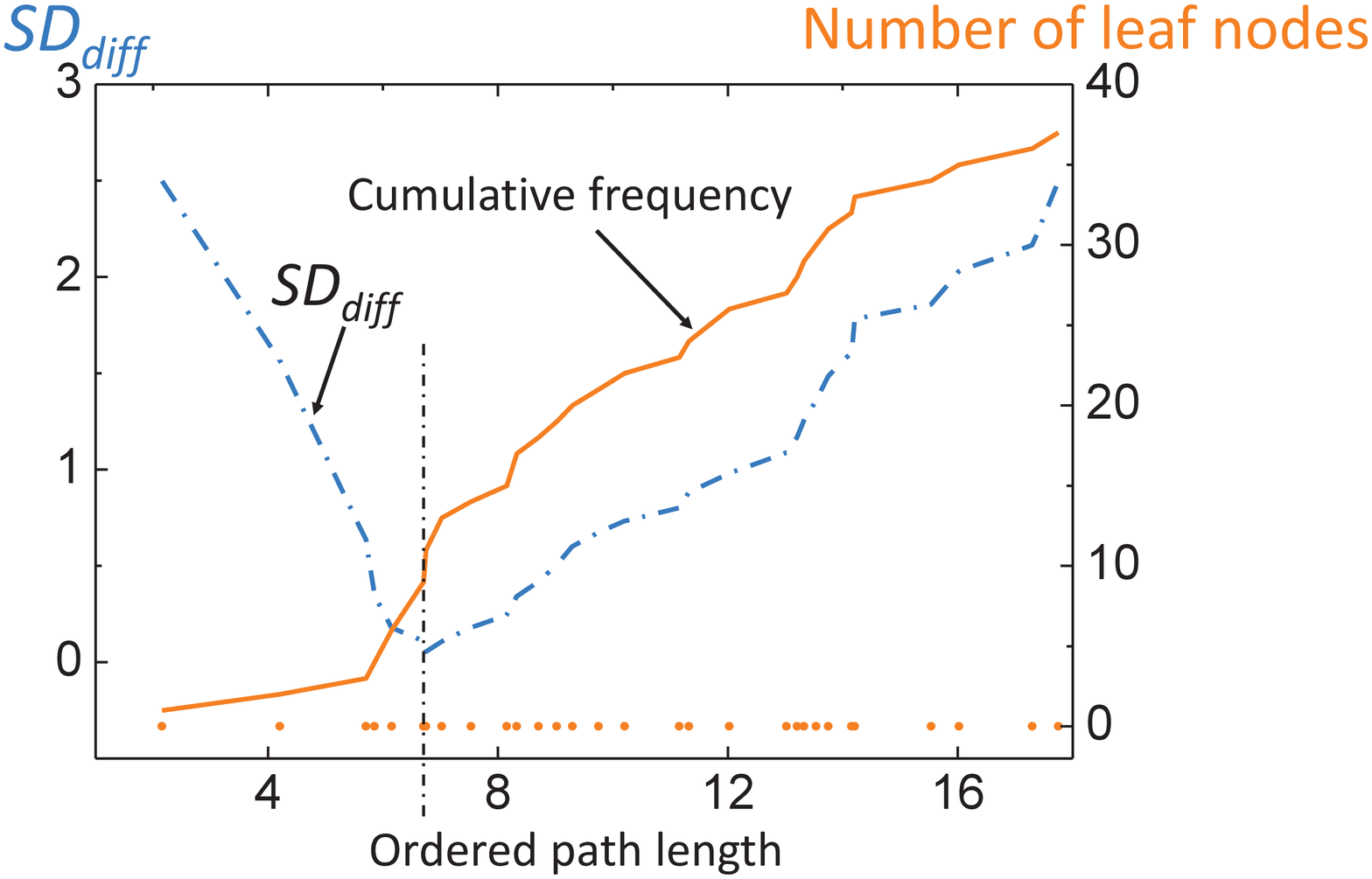,height=1.7in,width=3in}
  \caption{Determining path length threshold: Cumulative frequency and the corresponding $\emph{SD}_{\emph{diff}}$ curve, where the x-axis is the ordered path lengths from all regions in an iTree. The point which yields the minimum $\emph{SD}_{\emph{diff}}$ is chosen as the threshold to differentiate anomaly regions $A$ from normal regions $K$.}
 \label{deter}
\vspace{-3mm}
\end{figure}

Using a tree, Figure \ref{deter} shows an example of cumulative distribution for list $L$ and its $\emph{SD}_{\emph{diff}}$ ($=|\sigma(L^{r}) - \sigma(L^{l})|$) curve.
Note that the minimum $\emph{SD}_{\emph{diff}}$ point separates into two clear regions: anomaly and normal regions. 

Note that (i) because threshold $\hat{\tau}$ is determined automatically, no additional parameter is introduced; and (ii) this process does not require training data.

{\bf Construct ``outlying'' anomaly sub regions.}
\label{sec_outlying}
After $\hat{\tau}$ is determined,
 a ball $B$ is constructed using all training instances in every region $A$ of a tree, according to Definitions \ref{define:Anomaly of Known Class} and \ref{define:Outlying Anomalies}.

When balls $B$ have been built for all $A$ regions in every \emph{SENCTree}, the \emph{SENCForest} has the first function as an unsupervised detector and is ready to detect instances of emerging new classes.

A test instance which falls into $\cal A$ but outside $B$ is an ``outlying'' anomaly, i.e., an instance of an emerging new class.

{\bf Produce a classifier from a detector}
To incorporate the second function of being a classifier into \emph{SENCForest}, all we have to do is to record class distribution $F[j]$ in each region from $K$ and $B$ using the training subsample, where $F[j]$ denotes the number of class $j$ instances in a region. Note that this is the only step class labels are required.

Once the above training steps are completed, \emph{SENCForest} is ready to be deployed to a data stream.


\begin{algorithm}
\renewcommand{\algorithmicrequire}{\textbf{Input:}}  
\renewcommand{\algorithmicensure}{\textbf{Output:}}  
   \caption{Build \emph{SENCForest}}
   \label{top level}
\begin{algorithmic}[1]
  \REQUIRE ~ $D$ - input data, $z$ - number of trees, $\psi$ - subsample size.
  \ENSURE ~ \emph{SENCForest}
   \STATE {\bfseries initialize:} \emph{SENCForest} $\leftarrow$ \{\} 
    \FOR { $ i=1,\ldots,z $}
    \STATE $X_{i}$ $\leftarrow$ $sample(D, \psi)$
    \STATE \emph{SENCForest} $\leftarrow$ \emph{SENCForest} $\cup$ \emph{SENCTree}($X_{i}$)
   \ENDFOR
\end{algorithmic}
\end{algorithm}

 \begin{algorithm}
\renewcommand{\algorithmicrequire}{\textbf{Input:}}  
\renewcommand{\algorithmicensure}{\textbf{Output:}}  
   \caption{\emph{SENCTree}}
   \label{ACLTree}
   \begin{algorithmic}[1]
  \REQUIRE ~  $X$ - input data, $MinSize$ - minimum internal node size 
  \ENSURE ~ \emph{SENCTree}
  \IF {$|X| < MinSize$ }
   \STATE return LeafNode\{$|X|,F[\cdot],c,r$\}, as defined in Section~\ref{sec_enc}.
  \ELSE
  {
\STATE let $Q$ be a list of attributes in $X$
\STATE randomly select an attribute $q \in Q$
\STATE randomly select a split point $p$ from max and min values of attribute $q$ in $X$
\STATE $X_{L} \leftarrow$ $filter(X, q \le p)$
\STATE $X_{R} \leftarrow$ $filter(X, q > p)$
\STATE return inNode\{Left $ \leftarrow $ \emph{SENCTree}($X_{L}$),
\STATE \hspace{1.8cm} Right $\leftarrow$ \emph{SENCTree}($X_{R}$),
\STATE \hspace{1.8cm} SplittAtt $\leftarrow$ $q$,
\STATE \hspace{1.8cm} SplittValue $\leftarrow$ $p$ \},
  }
  \ENDIF
\end{algorithmic}
\end{algorithm}

\subsection{Deployment in data stream}
\label{sec_deployment}

Given a test instance $x$, \emph{SENCForest}($x$) produces a class label $y \in \{b_1,\dots,b_m,NewClass\}$, where $m$ is the number of known classes thus far and $NewClass$ is the label given for an emerging new class. Note that though \emph{SENCForest} can detect instances of any number of emerging new classes, they are grouped into one new class for the purpose of model update. We will focus on model update on one new class in one period (but multiple new classes could emerge in different periods of a data stream) for the rest of the paper. We discuss the issue of model update for multiple new classes in Section \ref{sec_results_condition}.

Algorithm \ref{Online Testing Process1} describes the testing process during the deployment of \emph{SENCForest} in a data stream.

 In line \ref{al3_line_output} of Algorithm \ref{Online Testing Process1}, \emph{SENCForest}($x$) outputs the majority class among all classes produced from $z$ trees. A tree outputs $NewClass$ if test instance $x$ falls into an $A$ region but outside the $B$ region; otherwise, it outputs the majority of class from
  \[ \argmax_{j \in \{b_1,\dots,b_m\} }  F[j]\]
\noindent
  where $F[j]$ is the class frequency for class $j$ recorded in the region  ($K$ or $B$) into which $x$ falls.


If \emph{SENCForest}($x$) outputs \emph{NewClass}, $x$ is placed in buffer $\mathcal{B}$ which stores the candidates of the previously unseen class (line \ref{al3_line_buffer}).  When the number of candidates has reached the buffer size, the candidates are used to update both the classifier and the detector (line \ref{al3_line_model_update}).
Once these updates are completed, the buffer is reset and the new model is ready for the next test instance in the data stream.

 \begin{algorithm}
\renewcommand{\algorithmicrequire}{\textbf{Input:}}  
\renewcommand{\algorithmicensure}{\textbf{Output:}}  
   \caption{Deploying \emph{SENCForest} in data stream}
   \label{Online Testing Process1}
\begin{algorithmic}[1]
  \REQUIRE ~  \emph{SENCForest}, $\mathcal{B}$ - buffer of size $s$
  \ENSURE ~  $y$ - class label for each $x$ in a data stream
  \WHILE {not end of data stream }{
  \FOR {each $x$}
  {
  \STATE $y$ $\leftarrow$ \emph{SENCForest}($x$)
  \label{al3_line_output}

 \IF {$y = NewClass$}  
\STATE $\mathcal{B}$ $\leftarrow$ $\mathcal{B}$ $\cup $ $\{ x \}$
\label{al3_line_buffer}
  \IF {$|\mathcal{B}|$ $\ge$ $s$}
    \STATE  Update (\emph{SENCForest}, $\mathcal{B}$)
    \label{al3_line_model_update}
   \STATE $\mathcal{B} \leftarrow$ NULL
   \STATE $ m \leftarrow m+1$
  \ENDIF
  \ENDIF
  \STATE Output $y \in \{b_1,\dots,b_m,NewClass\}$.

  }
  \ENDFOR
}
 \ENDWHILE
\end{algorithmic}
\end{algorithm}

\vspace{-4mm}
\begin{algorithm}
\renewcommand{\algorithmicrequire}{\textbf{Input:}}  
\renewcommand{\algorithmicensure}{\textbf{Output:}}  
\caption{Update \emph{SENCForest}}
   \label{updata1}
\begin{algorithmic}[1]
  \REQUIRE \emph{SENCForest} - existing model, $\mathcal{B}$ - input data
  \ENSURE a new model of \emph{SENCForest}
   \STATE {\bfseries initialize:} All instances in $\mathcal{B}$ are assigned a new class $b_{m+1}$
  \FOR { $ i=1,...,z $}
  \STATE $\mathcal{B}'$ $\leftarrow$ $sample(\mathcal{B}, \psi)$
\STATE Tree $\leftarrow$ \emph{SENCForest}.Tree[$i$]
  \FOR {$ j=1,...,$Tree.LeafNodeNumber}
  {
  \STATE $X'$ $\leftarrow$ instances of $\mathcal{B}'$ which fall into Tree.LeafNode$_{j}$
\IF{$|X'| > 0$ }
  \STATE $X$ $\leftarrow$ Pseudo instances from Tree.LeafNode$_{j}$
\STATE $X' \leftarrow X' \cup X $ \\
   \STATE Tree.LeafNode$_{j} \leftarrow$ \emph{SENCTree}($X'$)
   \label{al4_line_subtree}
\ENDIF
 }
 \ENDFOR
   \STATE recalculate $\hat{\tau}$ for Tree
  \STATE \emph{SENCForest}.Tree[$i$] $\leftarrow$ Tree
 \ENDFOR
\end{algorithmic}
\end{algorithm}

\subsection{Model Update}
\label{sec_Update}

\subsubsection{Growing Mechanism}

 \begin{figure}[h]
  \centering
  \epsfig{file=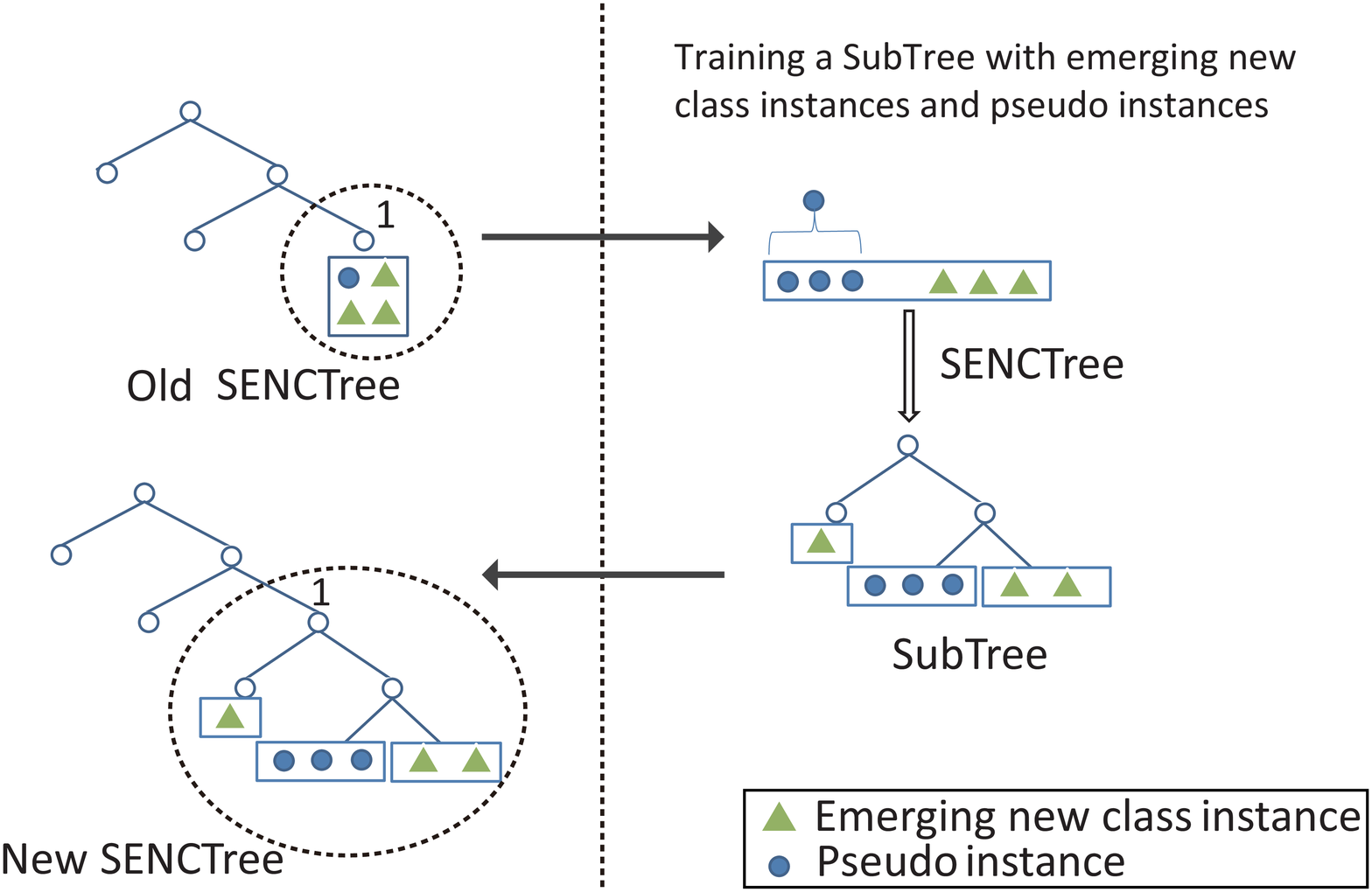,height=2.6in,width=3.3in}
 \vspace{-3mm}
  \caption{Replacing a leaf node with a trained subtree}
 \label{fig:Growing Mechanism}
\end{figure}

There are two growing mechanisms: one for growing a subtree in an \emph{SENCTree}, and the other for the growing multiple \emph{SENCForests}.

\noindent
{\bf Growing a subtree in an \emph{SENCTree}}. Updating \emph{SENCForest} with buffer $\mathcal{B}$ is a simple process of updating each leaf node in every tree using $\psi$ instances, randomly selected from $\mathcal{B}$. This is depicted in Algorithm~\ref{updata1}. The update at each node (line \ref{al4_line_subtree}) involves either a replacement with a newly grown subtree or a simple update of the class frequency to include the new class $b_{m+1}$.

If there are some instances which fall into a leaf node, a subtree needs to be grown as follows. As the previous training set is not stored, pseudo instances are generated for the leaf node which have the same attribute-values as centre $c$. The number of pseudo instances for each class $j$ is as recorded in $F[j]$. The combined set of pseudo instances $X$ and $X'$ (i.e., the subset of $\mathcal{B}'$ which falls into the same leaf node) is used as input to \emph{SENCTree} (line \ref{al4_line_subtree}). An example procedure is depicted in Figure \ref{fig:Growing Mechanism}. In the top left figure, we assume that some emerging new class instances (green triangle) fell into node 1 (there are three instances fell into in training process) in an \emph{SENCTree}. Then the combined set consists of pseudo instances and instances of the emerging new class. A new \emph{subTree} is built by using the combined set. Finally, in the bottom left figure, node 1 is replaced with this new \emph{subTree}. Every leaf node goes through the same process.

Note that the update process retains the original tree structure, and all pseudo instances in a leaf node will still be placed into a single leaf node of the newly grown \emph{subtree}. Thus, the predictions for the known classes are not altered in the model update process.

Once each tree has completed the model update, $\hat{\tau}$ is recalculated as described in Section \ref{sec_threhold}.

\noindent
{\bf Growing multiple \emph{SENCForests}}. When the number of classes in a \emph{SENCForest} reaches $\rho$, its \emph{SENCTrees} will stop growing for any emerging new class. A new \emph{SENCForest} is grown instead for the next $\rho$ emerging new classes.  This user-defined parameter is set based on the memory space available.

\subsubsection{Prediction using Multiple \emph{SENCForests}}
In a model with multiple \emph{SENCForests}, the final prediction is resolved as follows.
For a given $x$, \emph{SENCForest} $i$ yields prediction $y_{i}$ and probability
\begin{equation}
p_{i} = \frac{\mbox{Number of \emph{SENCTrees} predicting $y_i$}}{\mbox{Total number of \emph{SENCTrees}}}\nonumber
\end{equation}

The final prediction is $NewClass$ only if all \emph{SENCForests} predict $x$ as belonging to $NewClass$. Otherwise, the final prediction is the known class which has the highest $p_i$.
This procedure is given in Algorithm 5.

 \begin{algorithm}
\renewcommand{\algorithmicrequire}{\textbf{Input:}}  
\renewcommand{\algorithmicensure}{\textbf{Output:}}  
   \caption{Final Prediction from $E$ $\emph{SENCForests}$}
   \label{Voting Mechanism}
\begin{algorithmic}[1]
  \REQUIRE ~  $x$ - an instance in the data stream
  \ENSURE ~  $y_\imath$ - class label for $x$
  \FOR {$ i=1,...,E$}
  {
  \STATE $\left\langle y_{i}, p_i \right\rangle$ $\leftarrow$ $\emph{SENCForest}_{i}$($x$)
  \label{al5_line_output}
  }
   \ENDFOR
 \IF {$\forall_i\ \ y_i = NewClass$}  
  \STATE $\imath$ $\leftarrow$ $1$

  \ELSE
  {
    \STATE $L \leftarrow  \{i \in \{ 1,\dots,E \}\ |\ y_{i} \ne NewClass \}$
    \STATE  $\imath \leftarrow \mathop{\argmax}_{i \in L}\ p_{i}$
 }
 \ENDIF
  \STATE Output $y_\imath$

\end{algorithmic}
\end{algorithm}

\subsubsection{Retiring Mechanism}

A mechanism to retire \emph{SENCForest} is required as the data stream progresses. A \emph{SENCForest} is retired  under the following scenarios:
\begin{enumerate}
	\item When a \emph{SENCForest} is not used for predicting known classes for a certain period of time, it is eliminated for any future predictions. In other words, a \emph{SENCForest} outputs ``NewClass'' for a long time, this \emph{SENCForest} will be retired

\item In the event that the number of \emph{SENCForests} has reached the preset limit $\rho$ and no \emph{SENCForest} can be retired based on (1), then the least used \emph{SENCForest} in the last period is chosen to retire.
\end{enumerate}

The number of known class predictions is recorded for each \emph{SENCForest} in data stream. The one which has made the minimum number of predictions for known classes is identified to be the least used \emph{SENCForest}.

\begin{figure}[t]
  \centering
  \subfigure[Classification result in a data stream]{\epsfig{file=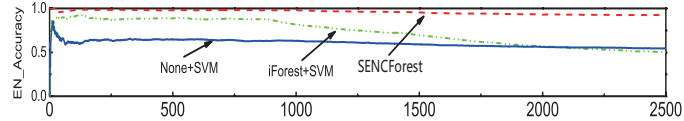,height=0.6in,width=3.5in}}
\subfigure[Emerging new class detection result of iForest+SVM]{\epsfig{file=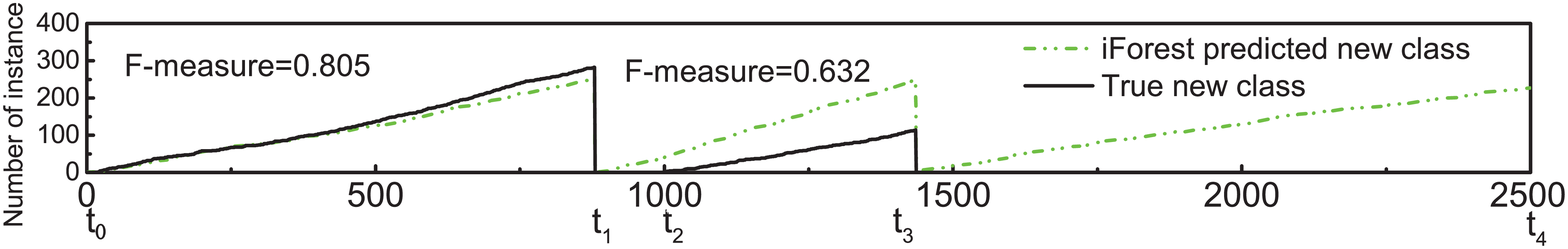,height=0.75in,width=3.5in}}
\subfigure[Emerging new class detection result of ENCiFer]{\epsfig{file=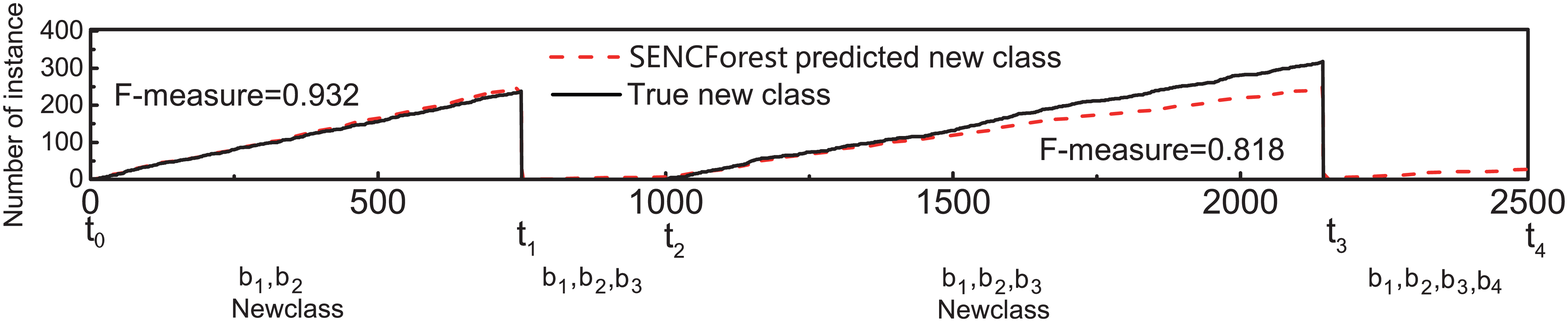,height=0.85in,width=3.5in}}
\vspace{-2mm}
  \caption{An example data stream on the KDDCUP 99 data set.
  The x-axis is the time steps in the data stream. The known classes at each duration ($t_i$ - $t_{i+1}$) are denoted as $b_1$,$b_2$,$b_3$, and $b_4$. The details of the two methods, iForest+SVM and None+SVM, are described in Table \ref{tbl_methods}. }
   \label{exampledatastream}
\end{figure}

\section{Experiment}
This section reports the empirical evaluation we have conducted to assess the performance of \emph{SENCForest} in comparison with several state-of-the-art methods.

\begin{figure}[t]
  \centering
  \subfigure[Classification result in a data stream]{\epsfig{file=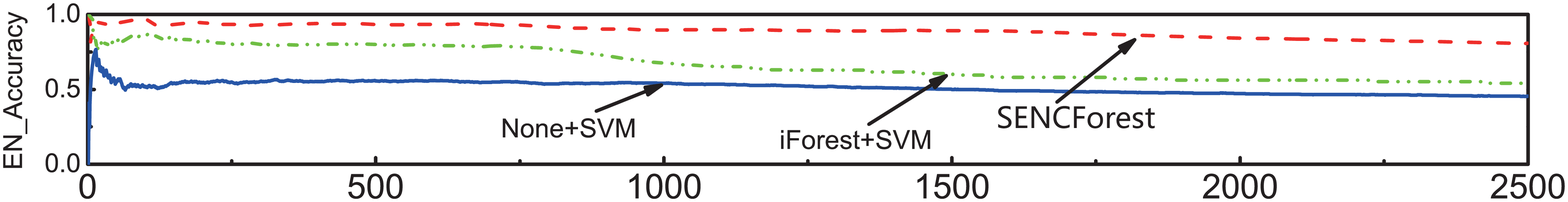,height=0.6in,width=3.5in}}
\subfigure[Emerging new class detection result of iForest+SVM]{\epsfig{file=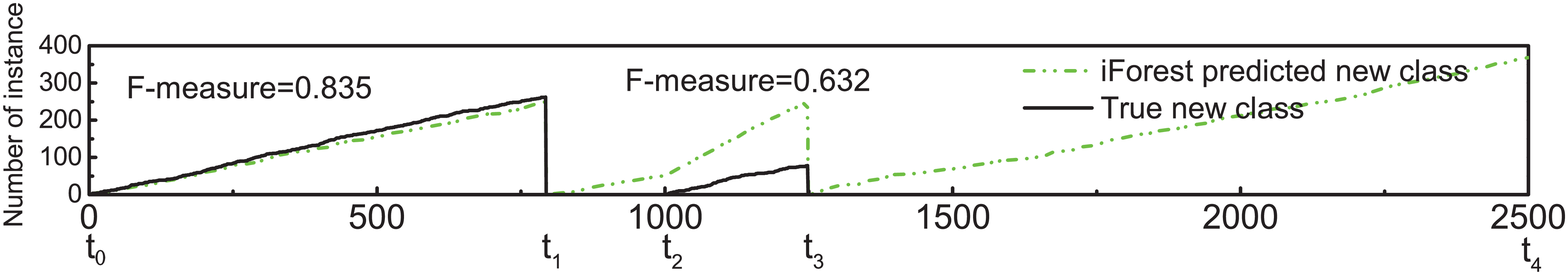,height=0.75in,width=3.5in}}
\subfigure[Emerging new class detection result of ENCiFer]{\epsfig{file=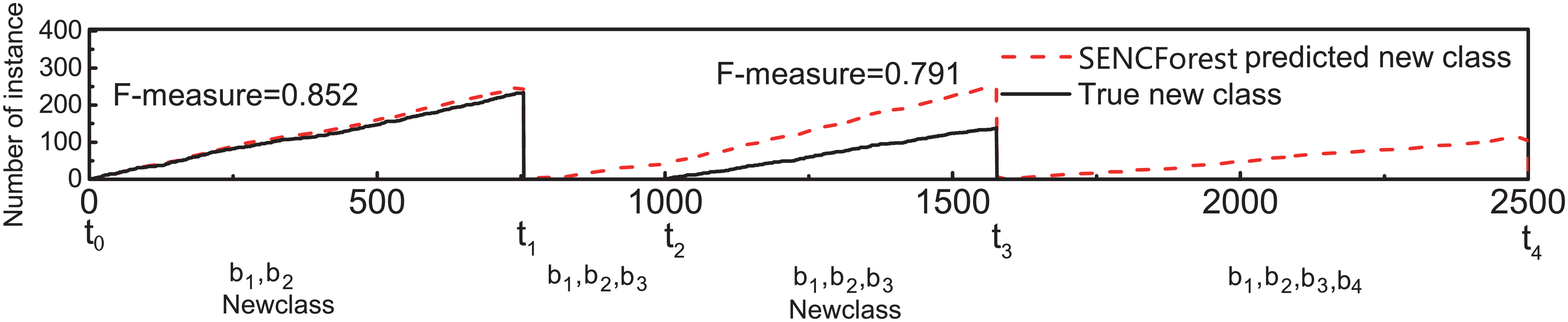,height=0.85in,width=3.5in}}
\vspace{-2mm}
  \caption{An example data stream on the MNIST data set.}
   \label{exampledatastream1}
\end{figure}

 \subsection{Experimental Setup}

{\bf Data Stream}:
To simulate emerging new classes in a data stream, we assume that an initial training set with two known classes are available to train the initial models. When the trained models are deployed at the beginning of a data stream, instances of the two known classes and an emerging new class appear in the first period of the data stream with uniform distribution. It is assumed that the method employed will update its models sometime within the first period. In the second period, instances of the three classes seen in the first period and another emerging new class appear with uniform distribution. Instances appear one at a time, and the deployed method is expected to make a prediction for each instance before processing the next, i.e., each instance is predicted as belonging to either an emerging new class or one of the known classes thus far.

No true class labels for all instances are available throughout the entire data stream.\footnote{This is a more stringent condition than previous studies (e.g., \cite{a5453372}) which assume that true labels are available for model update, after some time delay.} Model update is based on the instances of the emerging new class identified at the time the model update is triggered.

Figures \ref{exampledatastream} and \ref{exampledatastream1} show example data streams using the KDDCUP 99 data set and the MNIST data set. The class composition in the two distinct periods in the data stream are described as follows:

\begin{table}[h]
\centering
  \begin{tabular}{l |llc}
  &  & Known classes & New class\\ \hline
  First period:  $t_0$ - $t_2$ & $t_0$ - $t_1$ & $b_1$,$b_2$ & $\checkmark$\\
                 &  $t_1$ - $t_2$ & $b_1$,$b_2$,$b_3$ & $\times$\\ \hline
  Second period:  $t_2$ - $t_4$ & $t_2$ - $t_3$ & $b_1$,$b_2$,$b_3$ & $\checkmark$\\
                 &  $t_3$ - $t_4$ & $b_1$,$b_2$,$b_3$,$b_4$ & $\times$\\
   \end{tabular}
\end{table}

In the first period, all instances of the emerging new class identified by a method is placed in a buffer $\mathcal{B}$ of size $s$. When the buffer is full (marked as $t_1$), the method updates its model before processing the next instance. Note that $t_1$ differs for different methods as their detection rates for the new class are different, as shown in Figures \ref{exampledatastream}(b) and \ref{exampledatastream}(c) for iForest+SVM and \emph{SENCForest} (so as in Figures \ref{exampledatastream1}(b) and \ref{exampledatastream1}(c).) The buffer is reset to be empty when the model of a method has been updated. Note that after the model is updated, the new class in $t_0$ - $t_1$ becomes a known class $b_3$ of the updated model in $t_1$ - $t_2$, as shown in the table above.

Similarly, in the second period between $t_{2}$ and $t_4$, $t_3$ is the time when the buffer is full and the model of a method is updated for the second time. The new class in $t_2$ - $t_3$ becomes a known class $b_4$ of the updated model in $t_3$ - $t_4$.

\begin{figure}
  \centering
   \subfigure[]{\epsfig{file=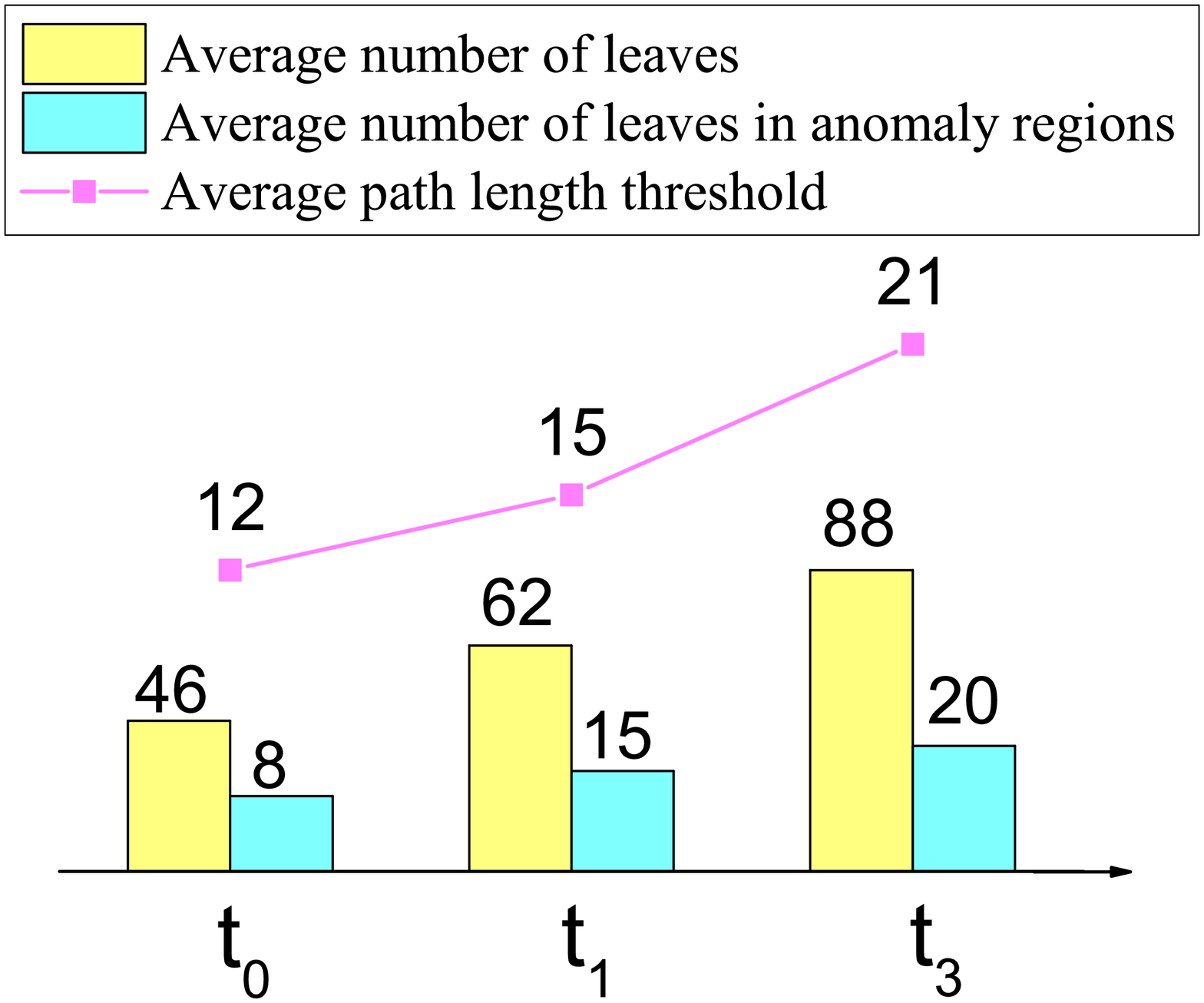,height=1.5in,width=1.6in}}
    \subfigure[]{\epsfig{file=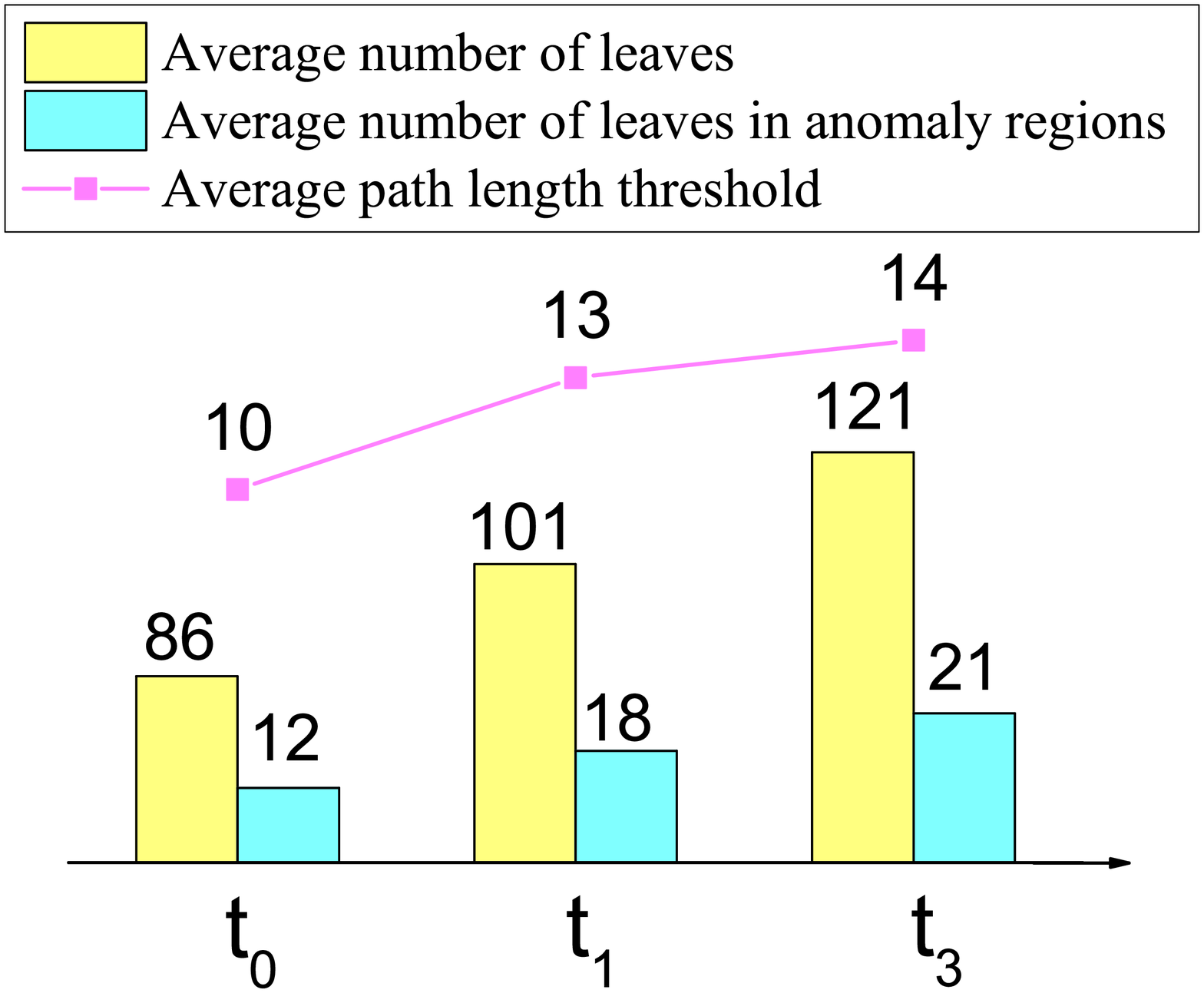,height=1.5in,width=1.6in}}
  \caption{Information of the evolving
  \emph{SENCForest} at three
different times in the data stream on (a) KDDCUP 99 data set; (b) MNIST data set.} 
 \label{fig:evoill}
\vspace{-3mm}
\end{figure}

Figure \ref{fig:evoill} shows the information of the evolving \emph{SENCForest} at three
different times in the data stream on two data sets.

\begin{table*}
\tiny
  \centering
  \caption{Methods used in the empirical evaluation. $D$ is the training set for the current models; $\mathcal{B}$ is the set of new class instances in the buffer and model update is triggered when the buffer is full. After each model update, $D \leftarrow D \cup \mathcal{B}$; and $D$ needs to be stored for the next model update for all methods, except \emph{SENCForest} and None+SVM.  $U$ is an additional set of unlabelled instances used by LACU-SVM only. In the experiments, the data size of $U$ is the total data size of $D$ and $\mathcal{B}$.}
  \label{tbl_methods}
  \begin{tabular}{|c|c|c|c|}
    \hline
   Method    & Detection         &    Classification & Model Update\\
       \hline
     LOF+SVM &  LOF              &  multi-class SVM & train new LOF and SVM with $D \cup \mathcal{B}$\\
        \hline
     1SVM+SVM &  one-class SVM  &  multi-class SVM & train new 1SVM  and SVM with $D \cup \mathcal{B}$\\
        \hline
     1R-SVM &  \multicolumn{2}{c|}{One-vs-rest SVM}  & train new 1R-SVM with $D \cup \mathcal{B}$\\
        \hline
     LACU-SVM & \multicolumn{2}{c|}{LACU-SVM}  & train new LACU-SVM with $D \cup \mathcal{B}$ and $U$\\
        \hline
          ECSMiner & \multicolumn{2}{c|}{ECSMiner} & train a new classifier in each fixed interval, assuming true labels are given\\
        \hline
     iForest+SVM &  iForest &  multi-class SVM  & train new iForest and SVM with $D \cup \mathcal{B}$ \\
        \hline
     \emph{SENCForest}+SVM &  \emph{SENCForest} &  multi-class SVM  & Update \emph{SENCForest} with $\mathcal{B}$ and train new SVM with $D \cup \mathcal{B}$ \\
        \hline

     None+SVM &  No detector &  multi-class SVM  & no model update \\
     \hline
     \emph{SENCForest} & \multicolumn{2}{c|}{\emph{SENCForest}} & Update \emph{SENCForest} with $\mathcal{B}$\\
    \hline
  \end{tabular}
\end{table*}

{\bf Evaluation measures}:
To evaluate the predictive accuracy of algorithms in the \emph{SENC} problem, we introduce EN\_Accuracy in a fixed window size. Let $N$ be the total number of instances in a window; $A_{n}$ be the total number of emerging class instances identified correctly; and $A_{o}$ be the total number of known class instances classified correctly,
\begin{equation*}
EN\_Accuracy = \frac{A_{n}+A_{o}}{N}
\end{equation*}

Figures \ref{exampledatastream}(a)  and \ref{exampledatastream1}(a) show examples of EN\_Accuracy results of three methods in a data stream.

To evaluate the accuracy of new class detection, we compute F-measure 
in $t_0$ - $t_1$ and $t_2$ - $t_3$ to measure the detection performance in these two durations. This measure produces a combined effect of precision (P) and recall (R) of the detection performance. F-measure = 1 if a detector identifies all instances of emerging new class with no false positives.
\begin{equation*}
  F\texttt{-}measure = \frac{2*P*R}{P+R}
\end{equation*}

The cumulative numbers of instances of the true and  predicted new class are also plotted in four consecutive durations. In $t_{1}$-$t_4$, it shows that both methods make some false positives resulting in more instances predicted as belonging to the new class than it actually has. The F-measures achieved by each detection method in $t_{0}$-$t_1$ and  $t_{2}$-$t_3$ are shown in Figures \ref{exampledatastream}(b) \& \ref{exampledatastream}(c) and Figures \ref{exampledatastream1}(b) \& \ref{exampledatastream1}(c).
In this example, \emph{SENCForest} performs better than iForest+SVM because it has better F-measure, fewer false positives and higher EN\_Accuracy.

In the experiments reported in Section \ref{sec_results}, the difference in performance between two methods is considered to be significance on paired t-tests at 95\% significance level in our paper

{\bf Contenders}:
The complete list of the methods used for new class detection, classification and model update methods is shown in Table \ref{tbl_methods}. As some of these methods can act as a new class detector only, a state-of-the-art classifier, i.e., multi-class SVM \cite{CC01a}, is employed to classify instances of known classes.
Note that three types of information, additional to that was provided to \emph{SENCForest}, are required for other methods. First, true labels must be provided at each model update.  Otherwise, no models could be updated. ECSMiner assumes that true labels are given at the end of a fixed interval ($T_l$) in order to update model. Other existing methods requires all instances in $\mathcal{B}$ must be given the true labels. Second, LACU-SVM needs to have additional unlabelled data before training at each model update.
Third, the initial training set must be stored and incorporated at each model update. \emph{SENCForest} is the only method which does not require (i) true labels during the entire data stream after training, (ii) to store the initial training set, and (iii) unlabelled training set.


A brief description of each of the methods used in the experiment is given as follows:

\begin{enumerate}
  \item {\bfseries LOF} or Local Outlier Factor \cite{DBLP:conf/sigmod/BreunigKNS00} is a density-based anomaly detector which employs k-nearest neighbour procedure to estimate density.
  \item {\bfseries One-class SVM} \cite{Scholkopf:2001:ESH:1119748.1119749} is a state-of-the-art outlier detector \cite{ma2003time} which learns from normal instances only. It computes a binary function to capture regions in input space where the probability density lives.
  \item {\bfseries One-vs-rest SVM} is a scheme for multi-class classification \cite{Rifkin:2004:DOC:1005332.1005336} where a two-class SVM $f_k(\cdot)$ is built for each class. In the original One-vs-rest SVM, a test instance $x$ is predicted as belonging to class $k$ if $f_k(\cdot)$ produces the highest confidence. To adapt One-vs-rest SVM to predict the emerging new class, the classifier produces a classification prediction only if $\max_{k} f_{k}(x) > 0$; otherwise $x$ is predicted as belonging to the emerging new class.
  \item {\bfseries LACU-SVM} \cite{DBLP:conf/aaai/DaYZ14} is a semi-supervised learner which modifies a previously trained model by considering the structure presented in the unlabelled data so that the misclassification risks among the known classes as well as between the new and the known classes are minimized simultaneously. It produces a classifier which predicts one of the known classes or the new class. This method also trains $k$ binary classifiers $f_k(\cdot)$ for each known class. Like One-vs-rest SVM, LACU-SVM makes a prediction for the known class if  $\max_{k} f_{k}(x) > 0$; otherwise $x$ is predicted as belonging to the emerging new class.
   \item {\bfseries ECSMiner} \cite{a5453372} is an algorithm for novel class detection and classification. It employs the clusters identified by k-means to detect novel classes: instances which are not within the boundaries of any clusters are treated as novel class candidates and placed in a buffer, then a new measure is defined to decide whether they are emerging new classes. K nearest neighbor is used as the classifier to make predictions for instances of known classes. Model update can only occurs if true labels are available within some fixed duration.

  \item  {\bfseries iForest} \cite{liu2008isolation} is an unsupervised anomaly detector which builds a model to isolate each training instance from the rest of the training set.

\end{enumerate}

In the experiments, all methods were executed in the MATLAB environment. The following implementations are used: SVM in the LIBSVM package \cite{CC01a}; LACU-SVM and iForest were the codes as released by the corresponding authors; and LOF is in the outlier detection toolbox.\footnote{https://goker.wordpress.com/2011/12/30/outlier-detection-toolbox-in-matlab/} The ECSMiner code is completed based on the authors' paper \cite{a5453372}. We set the max size of each tree to 300, which avoids to the worst case that growing infinitely by random partition. The parameter settings used for these algorithms are provided in Table \ref{parametersettings} in Appendix A.

{\bf Data sets}: Five data sets are used to assess the performance of all methods , including Synthetic, KDDCup 99\footnote{http://kdd.ics.uci.edu/databases/kddcup99/kddcup99.html}, Forest Cover\footnote{https://kdd.ics.uci.edu/databases/covertype/covertype.data.html}, MHAR and MNIST\footnote{http://cis.jhu.edu/~sachin/digit/digit.html}. For KDDCup 99 data set, we use the four largest classes, i.e., normal, neptune, smurf and back. For Forest Cover data set, we use 10 attributes, and all binary attributes are removed. A description for Synthetic and MHAR data sets are provided in Appendix B. A summary of the data characteristics is provided in Table \ref{tbl_data}.

\begin{table}[h]
  \centering
  \caption{A summary of data sets used in the experiments.}
  \label{tbl_data}
  \begin{tabular}{|c|c|c|c|}
    \hline
      Data set & \#classes & \#attributes   \\
       \hline
     Synthetic &  4  &  2 \\
        \hline
    KDDCup 99& 4 & 41 \\
        \hline
    Forest Cover &  7  & 10  \\
        \hline
     MHAR  &  6  &  561 \\
    \hline
     MNIST&  10  &  784 \\
        \hline
  \end{tabular}
\end{table}

{\bf Simulation}: In the following experiment, each data set is used to simulate a data stream over ten trials. In each trial, the initial training set has two classes, and the emerging new class in each period is a class different from the known classes. These classes are randomly selected from the available classes. The instances in the initial training set and the data sequence in the data stream are randomly selected from the given data set, but following uniform class distribution. For all real-world data sets, the data size of the initial training set $D$ is 500 per class; the buffer size $|\mathcal{B}| =250$; and the total number of instances which have appeared in the data stream at the end of the first period at $t_{2}$ is 1000; and the second period ($t_2$ - $t_4$) has a total of 1500 instances. As we can afford to generate more data in the synthetic data set, $D$, $\mathcal{B}$, and the data size at each period are double to examine the effect of larger data sizes. The average result of ten trials is reported.

 The following sections will give related evaluation results. Section \ref{sec_results} describes the empirical evaluation under the condition that no true  labels are available after the data stream has started. Section \ref{sec_results_long} reports results under the long streams situation. Section \ref{sec_results_condition} describes using
 \emph{SENCForest} under the condition that emerging multiple new classes in a period.

\subsection{Empirical results}
\label{sec_results}

\begin{figure}
  \centering
  \subfigure[Synthetic]{\epsfig{file=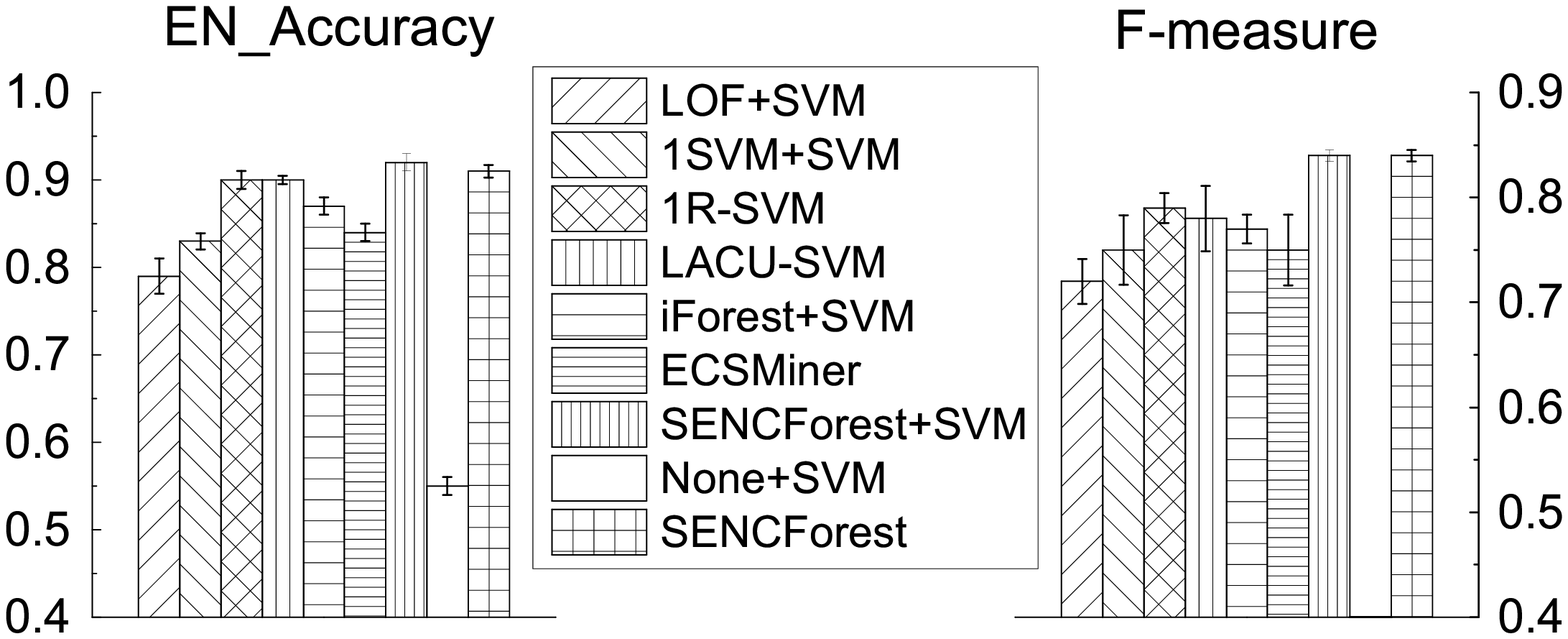,height=1.3in,width=3in}}
  \subfigure[KDD Cup 99]{\epsfig{file=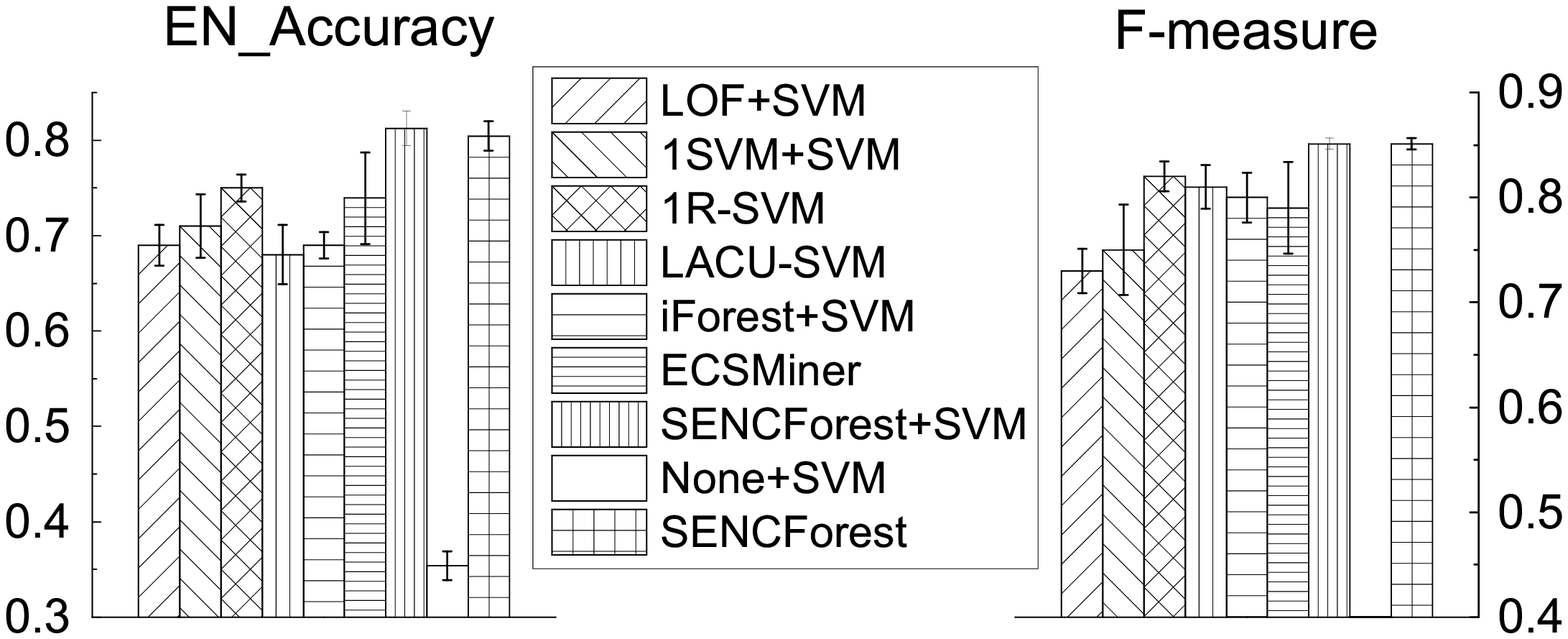,height=1.3in,width=3in}}
\subfigure[Forest Cover]{\epsfig{file=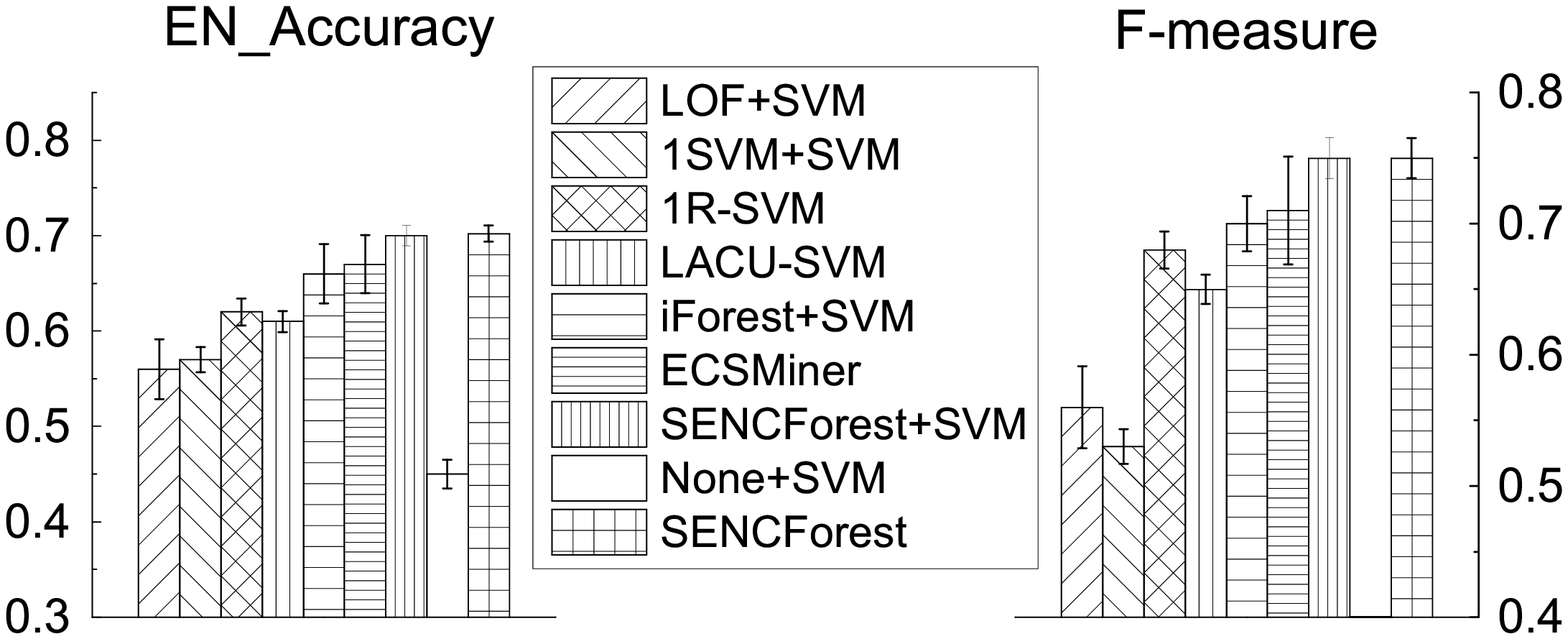,height=1.3in,width=3in}}
\subfigure[MHAR]{\epsfig{file=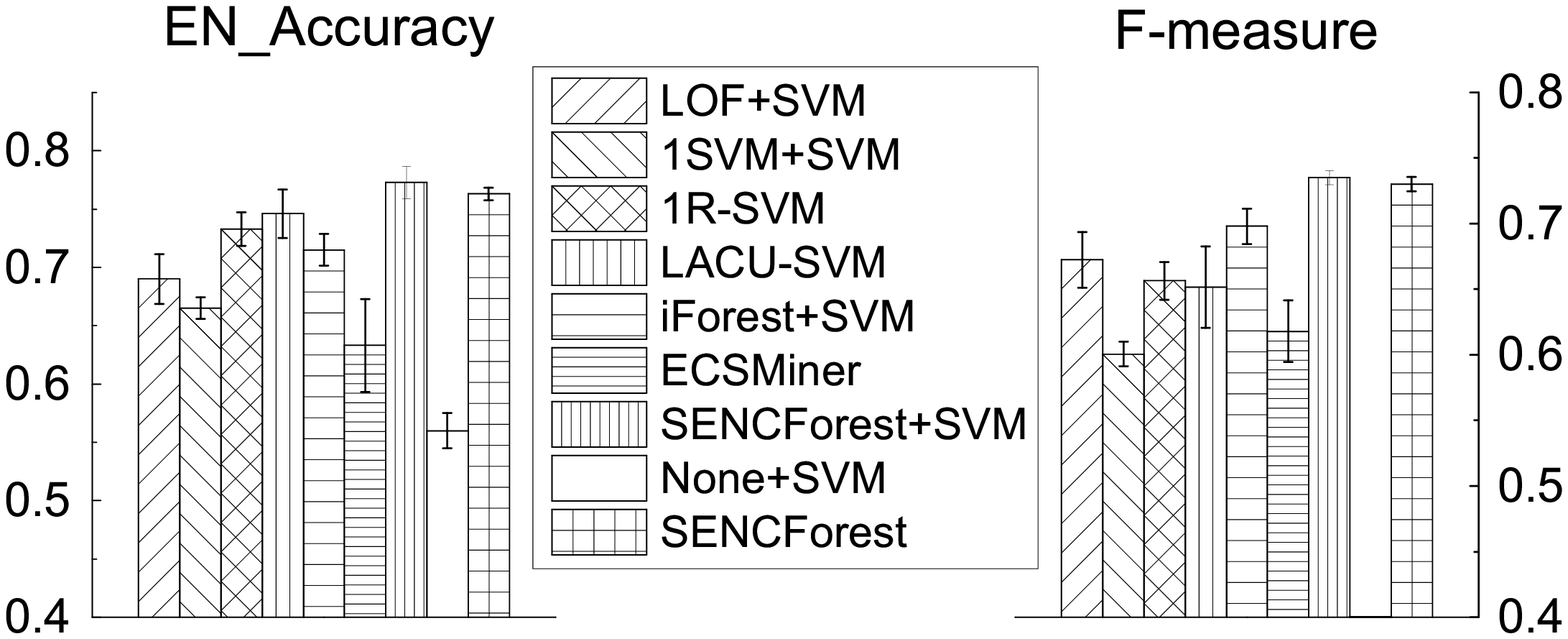,height=1.3in,width=3in}}
\subfigure[MNIST]{\epsfig{file=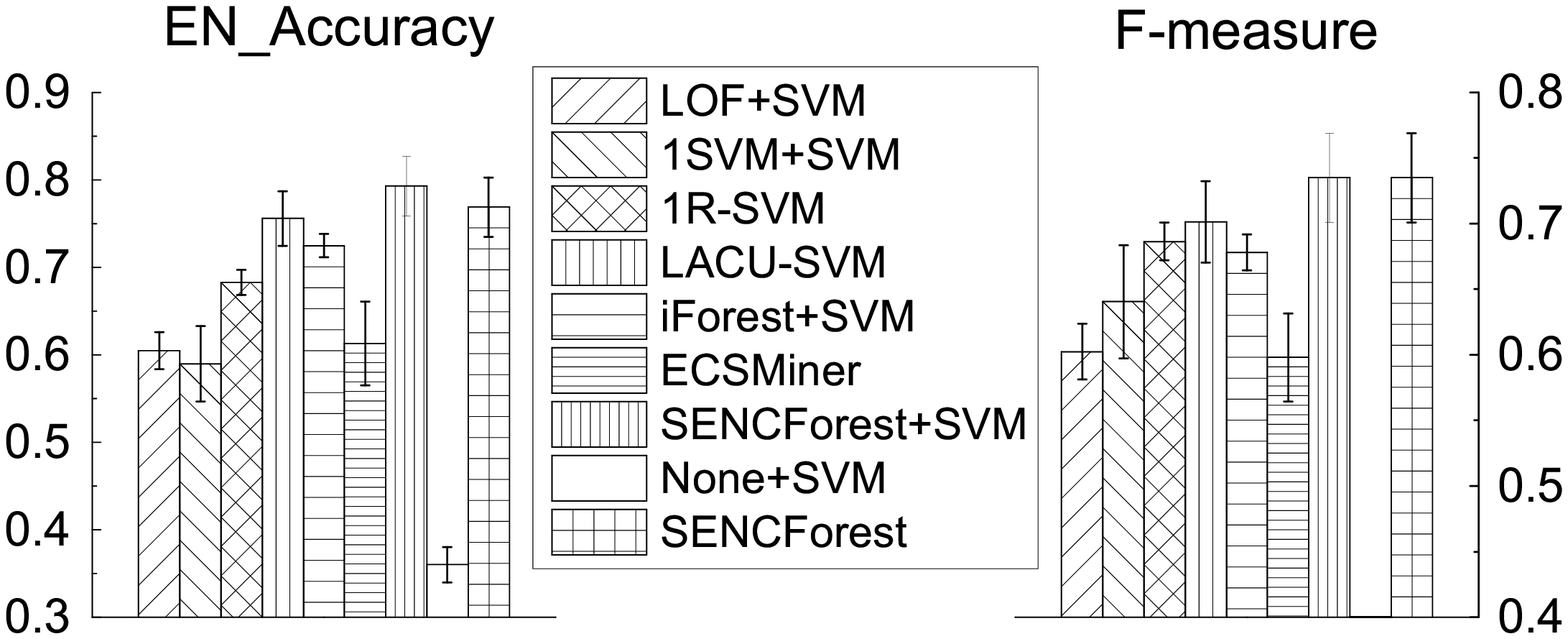,height=1.3in,width=3.3in}}
\caption{Average result over ten trials. In each trial, the following is computed: Average F-measure in $t_0$ - $t_1$ and $t_2$ - $t_3$; and average accuracy over the entire duration from $t_0$ to $t_4$. Two standard errors over ten trials are shown as the error bar. Note that \emph{SENCForest} and \emph{SENCForest}+SVM are using the same detector to detect emerging new class; thus they have the equivalent F-measure result.}
   \label{fig_result}
   \vspace{-3mm}
\end{figure}

The results for the five data sets are shown in Figure \ref{fig_result}.

In terms of new class detection, \emph{SENCForest} produced the highest F-measure in all data sets. Recall that \emph{SENCForest}+SVM uses \emph{SENCForest} only for new class detection; thus both \emph{SENCForest} and \emph{SENCForest}+SVM have the same F-measure performance.

The closest contenders are LACU-SVM and 1R-SVM, each had the second or third highest F-measure in three data sets. \emph{SENCForest} was significantly better than all contenders, except in MNIST (wrt LACU-SVM) and Forest Cover (wrt ECSMiner).


In terms of EN\_Accuracy, \emph{SENCForest} and \emph{SENCForest}+SVM produced the highest performance in all data sets. This result shows that (i) the accurate detection of emerging new class leads directly to high classification accuracy; and (ii) \emph{SENCForest} as a classifier is competitive to SVM. LACU-SVM was the closest contender which had the second highest accuracy in three data sets. Beside \emph{SENCForest}+SVM, \emph{SENCForest} performed significantly better than the other contenders in three data sets. The two exceptions are wrt to LACU-SVM (in MNIST and Synthetic) and 1R-SVM (in Synthetic).

An analysis is provided below:
\begin{itemize}
\item LOF and one-class SVM: the poor detection performance of these two methods wrt to iForest is likely to be due to the parameter search, i.e., a search for a wider range of values may improve their performance. However, such search is a computationally expensive process, and this makes them unsuitable for data stream applications.
\item iForest performed worse than \emph{SENCForest} in all data sets, and the differences were significance in four data sets. This shows that an unsupervised anomaly detector can be successfully used in the \emph{SENC} problem if anomaly regions are reshaped (as described in Sections \ref{sec_threhold}) to detect emerging new classes.

\item While One-vs-rest SVM performed reasonably well in classification, it is not a good choice for detection of emerging new classes, in comparison with \emph{SENCForest}.
\item LACU-SVM is the only method which requires additional unlabelled instances in training the initial model and in every model update. While obtaining unlabelled instances may not be a problem in real applications, it is important to note that its detection performance is highly depended on the existence of a new class in the set of unlabelled instances. Insufficient instances of the new class will severely limit LACU-SVM's ability to detect the new class. In the experiment, LACU-SVM was provided a set of unlabelled instances in $t_0$, $t_1$ and $t_3$, in addition to those instances in the initial training set and the buffer, in order to update its model. This additional data set was not available to all other methods.
Despite this additional training information, LACU-SVM still performed significantly worse than \emph{SENCForest} in four data sets
in terms of F-measure.
\item ECSMiner is the only algorithm which was provided with true labels in order to train a new classifier in each fixed interval, which occurs more often than at each model update, over the entire data stream. Despite this advantage, it still performed significantly worse than \emph{SENCForest} in four out of five data sets in both measures.\footnote{ECSMiner \cite{a5453372} had employed the KDD CUP 99 and Forest Cover datasets in their evaluation. Our ECSMiner results are compatible with theirs in these two datasets. However, ECSMiner performed poorly in the other three datasets.}

\item The result of None+SVM clearly shows that not using a detector is not an option in the \emph{SENC} problem.
\item \emph{SENCForest} is the best choice detector and a competitive classifier in the \emph{SENC} problem. While it is possible that a more sophisticated classifier may yield a higher accuracy in classifying known classes, it often comes at a high computational cost in an extensive parameter search.

\item While using SVM, in addition to \emph{SENCForest}, could potentially produce a better accuracy than that from \emph{SENCForest} alone, this comes with a computational cost which is usually too expensive in the data streams context. Note that to achieve the performance of \emph{SENCForest}+SVM presented in Figure \ref{fig_result}, it needs to store all instances thus far, which is impossible in data streams. In contrast, \emph{SENCForest} achieves comparable result as \emph{SENCForest}+SVM without the need to store any data.
\end{itemize}

\subsection{\emph{SENCForest} in long data streams}
 \label{sec_results_long}

 The aims of this section are to examine the ability of \emph{SENCForest} to (i) maintain good performance using limited memory in long data streams; and (ii) make use of true class labels when they are available.

We simulate a long data stream using the MNIST data set. This stream has twelve emerging new classes\footnote{Classes are reused in the simulation when they are no more in use in the current period. Because this simulation needs a number of classes, that is why only the MNIST dataset, out of the five datasets, can be used in the long stream simulation.}. The initial training data set has 2 classes, and every subsequent period has 1000 instances from one emerging new class and two known classes. The maximum number of classes which can be handled by each \emph{SENCForest} is set to 3. Other settings are the same as used in the last section. In addition, true class labels are assumed to be available in $Q$ percentage of instances in the buffer before a model update. \emph{SENCForest} with $Q=0$\%, $50$\% and $100$\% are compared with LACU-SVM in the experiment.  Recall that, as in the previous experiment, LACU-SVM is given 100\% true labels at each model update and an additional set of unlabelled instances; and ECSMiner is also provided with 100\% true labels at each model update.

\begin{figure}
  \centering
  \epsfig{file=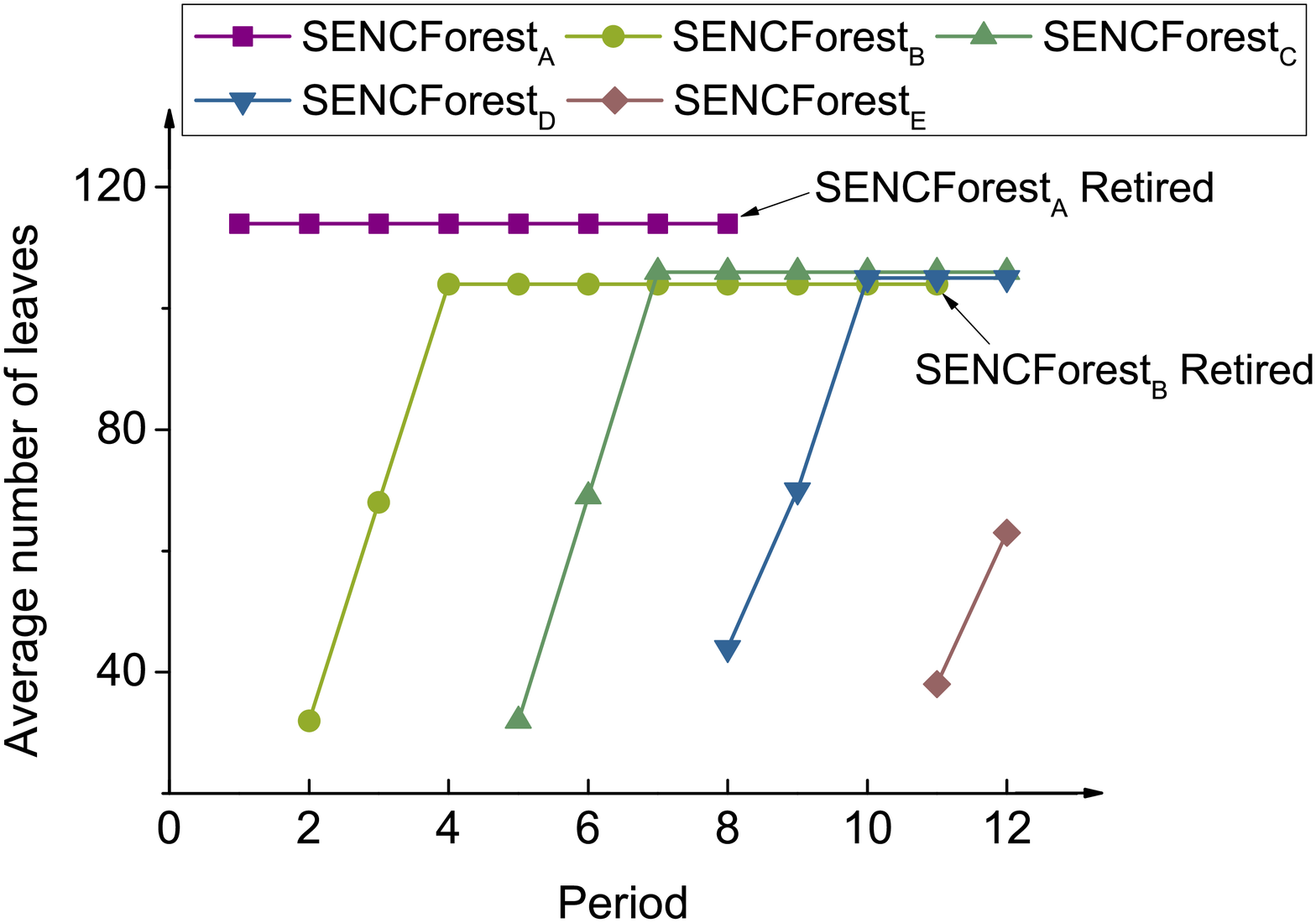,height=1.7in,width=2.3in}
  \caption{Average number of leaves of each evolving \emph{SENCForest} at the start of each time period.}
 \label{fig:retire}
\vspace{-3mm}
\end{figure}

 Figure \ref{fig:retire} shows the average number of leaves of each \emph{SENCForest} at the start of each time period. Note that a new \emph{SENCForest} was produced at periods 2, 5, 8, 11, and the first two \emph{SENCForests}, $A$ and $B$, were retired at periods 8 and 11,  respectively. Table \ref{evolving1} shows the further information  about \emph{SENCForests}($Q=0$\%) at the start of each time period. The first three rows provide the overall information; and the last three lines show the detailed information of the only evolving \emph{SENCForest} at each time period, e.g., periods 2, 3, 4 for $\emph{SENCForest}_B$, periods 5, 6, 7 for $\emph{SENCForest}_C$ and so on. Note that the number of leaves in anomaly regions may decrease as \emph{SENCForest} grows. This happens when instances of new classes fall into few leaves only.

The number of \emph{SENCForests} is maintained at a preset memory limit through retiring not-in-use \emph{SENCForests}. Note that the model size is constrained within the set limit of three SENCForests which allows the proposed method to deal with infinite data streams. In contrast, LACU-SVM continues to demand larger and larger memory size to accommodate larger training set size as the stream progresses.

\begin{figure}
  \centering
   \subfigure[]{\epsfig{file=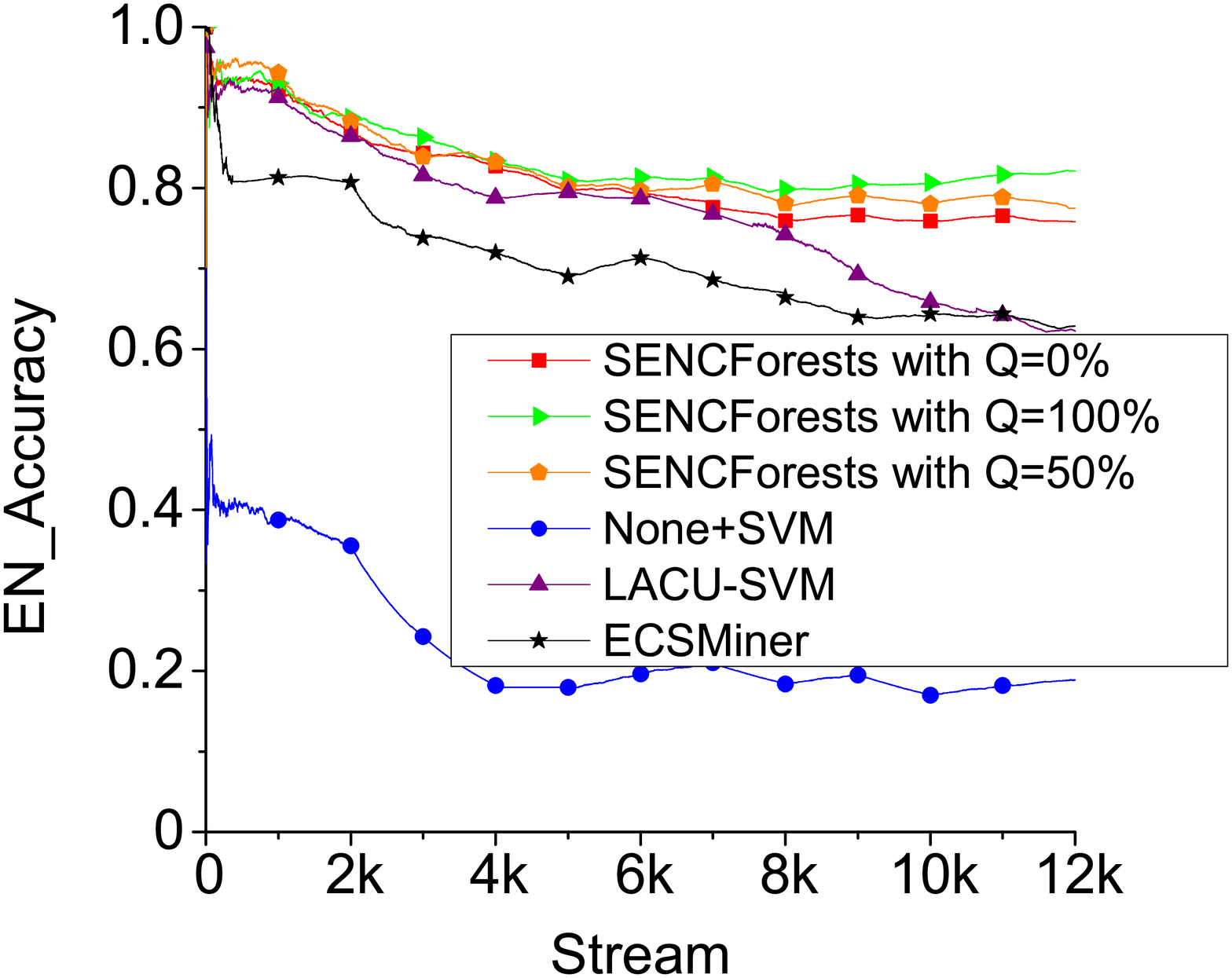,height=1.5in,width=1.6in}}
    \subfigure[]{\epsfig{file=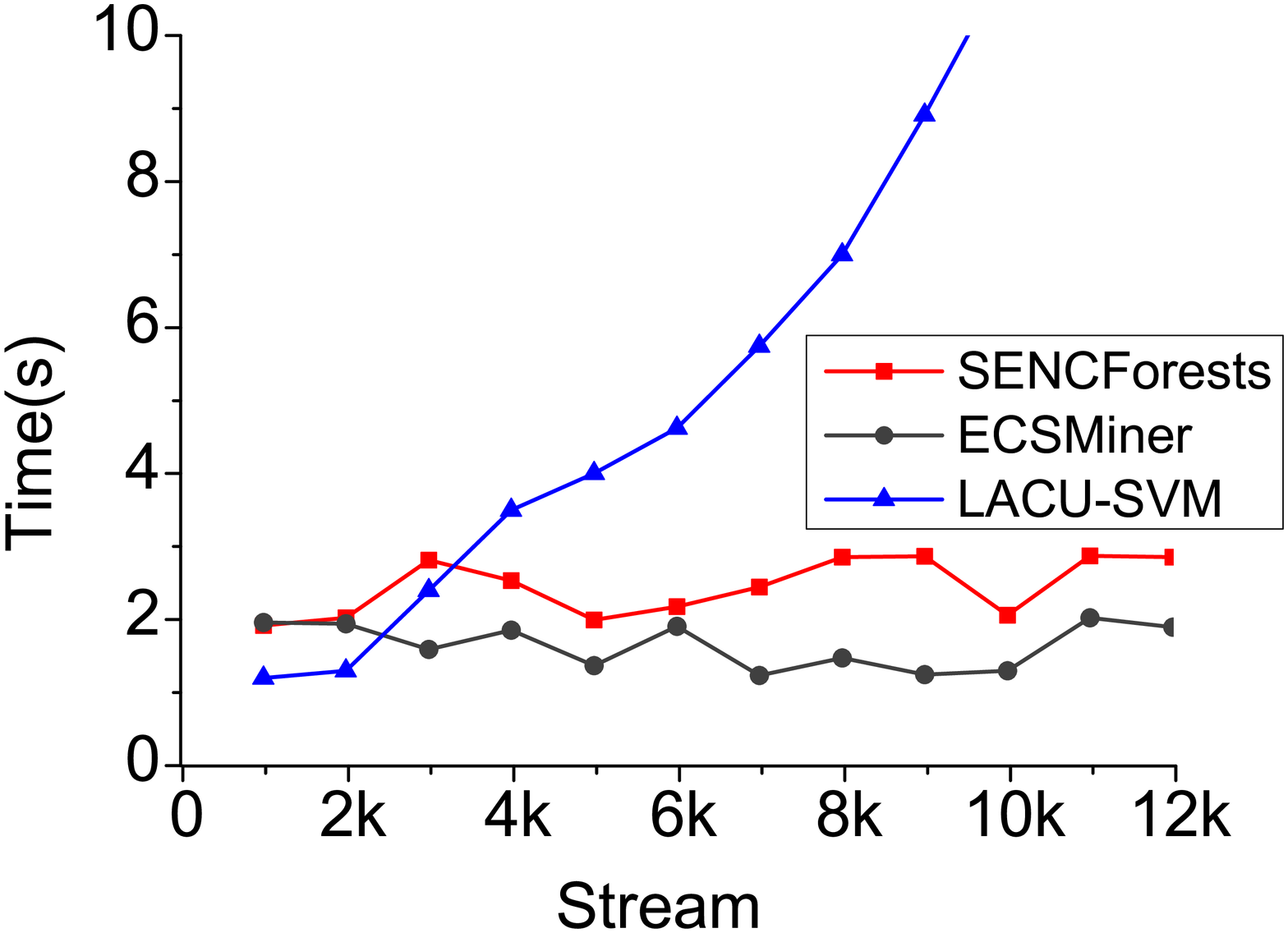,height=1.5in,width=1.6in}}
  \caption{(a) Result of long data stream in the MNIST data set; the bar chart on the bottom shows \emph{SENCForests} with $Q=0$\%, the number of \emph{SENCForests} and the number of  retired \emph{SENCForests} at each period. (b) The time spent (in seconds) to do model update in each period.} 
 \label{fig:long2}
\vspace{-3mm}
\end{figure}

  The result in Figure \ref{fig:long2} (a) shows that \emph{SENCForest} with $Q=0$\% maintains good predictive accuracy over the long stream.  \emph{SENCForest} is able to make use of true class labels to improve its performance along the stream. The extent of the improvement increases as $Q$ increases. In contrast, the predictive accuracy of ECSMiner and LACU-SVM continued to decrease as the stream progressed.

  As a result, as shown in Figure \ref{fig:long2}(b), its training time continued to grow as the stream progressed. ECSMiner has the least model update time,
because k-nearest neighbor is as base learner, which only spends time in building the clusters in buffer. But, that means it needs to save the cluster summary of each cluster into memory.




\subsection{Multiple new classes in a period}
\label{sec_results_condition}

\begin{table*}
 \tiny
  \centering
  \caption{Information of each evolving \emph{SENCForest} (marked with $^*$)  at the start of each time period on the simulated data stream using MNIST. Note that only the latest \emph{SENCForest} at any time period is evolving or growing; and all earlier built \emph{SENCForests} (if any) have stopped growing. The subscript indicates the latest \emph{SENCForest} shown in Figure \ref{fig:retire}.}
  \label{evolving1}
  \begin{tabular}{|c|c|c|c|c|c|c|c|c|c|c|c|c|}
    \hline
Period & 1&2& 3&4& 5&6& 7&8& 9&10& 11&12  \\
    \hline
Number of known classes & 3&4& 5&6& 7&8& 9&7& 8&9& 7&8  \\
        \hline
Number of \emph{SENCForests}& 1&2& 2&2& 3&3& 3&3& 3&3& 3&3  \\
        \hline
Number of retired \emph{SENCForests}& 0&0& 0&0& 0&0& 0&1& 0&0& 1&0  \\
    \hline
Average number of leaves$^*$& 114$_A$&32$_B$& 68$_B$ &104$_B$ & 32$_C$&69$_C$&106$_C$& 44$_D$&70$_D$& 105$_D$& 38$_E$& 63$_E$ \\
    \hline
Average number of leaves in anomaly regions$^*$& 14$_A$&11$_B$& 15$_B$&12$_B$& 9$_C$&12$_C$& 13$_C$&14$_D$& 13$_D$&14$_D$& 16$_E$&15$_E$  \\
    \hline
Average path length threshold$^*$& 16$_A$&8$_B$ &16$_B$& 17$_B$&10$_C$& 17$_C$&18$_C$& 11$_D$&17$_D$& 20$_D$& 11$_E$ & 15$_E$  \\
    \hline
  \end{tabular}

\end{table*}

\begin{figure}
  \centering
  \epsfig{file=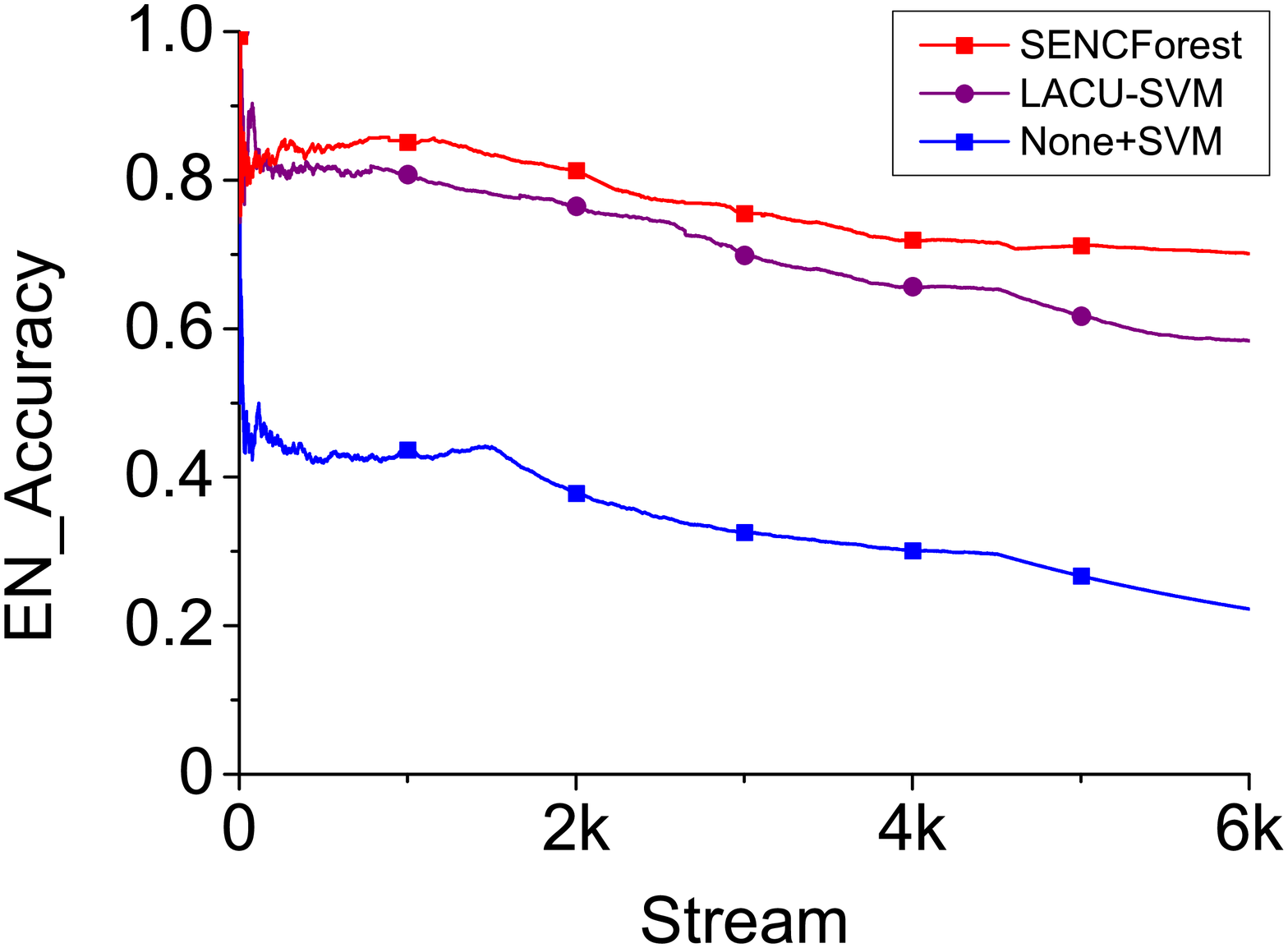,height=1.8in,width=2.2in}
  \caption{Result of two emerging new classes in each period.}
 \label{fig:long}
\vspace{-3mm}
\end{figure}

The emergence of multiple new classes in a period is a challenge in the \emph{SENC} problem.
Although \emph{SENCForest} is designed to deal with one emerging new class in each period, it can still perform well by treating these emerging classes in a period as a single new class.
Figure \ref{fig:long} shows that \emph{SENCForest} performs as well when every period has two emerging new classes. In this stream, there are three periods; each period has 2000 instances and 4 classes (i.e., two emerging new classes and two known classes).

In the event that it is important to identify each class in each period, a clustering algorithm\cite{DBLP:books/crc/aggarwal13/Aggarwal13a} can be used to achieve this aim before proceeding to do the model update.

%

\section{Conclusions and future work}

This paper contributes to decompose the \emph{SENC} problem into three sub problems and posits that the ability to tackle the first sub problem of detecting emerging new classes effectively is crucial for the whole problem. The difficulty of the \emph{SENC} problem is highlighted by the inability of existing methods to solve it satisfactorily.

We show that the unsupervised-anomaly-detection-focused approach, coupled with an integrated method using completely random trees, provides a complete solution for the entire \emph{SENC} problem. The current classification-focused approach has failed to provide one thus far.

The strength of \emph{SENCForest} is its ability to detect new class with high accuracy. The use of an unsupervised anomaly detector, incorporated with the new ability to differentiate between anomalies of known classes and instances of new classes, underlines the source of the strength. Existing supervised and semi-supervised methods are unable to achieve the same level of detection accuracy because the focus was on the second sub problem: classification, rather than the first sub problem: emerging new class detection.

The fact that the unsupervised learner consists of completely random trees facilitate the use of a common core which can be converted to an effective classifier with ease. The common core also makes model updates in data streams to be a simple model adjustment, rather than training a completely new model as in most existing methods. Like in previous work, we show that the completely random trees are a classifier competitive to state-of-the-art classifiers, especially in the data stream context which demands fast model update and classification time.

Our empirical evaluation shows that \emph{SENCForest} outperforms eight existing methods, despite the fact that it was not given the true class labels in the entire data stream; and other methods were given the true class labels at each model update. In addition, it works effectively in long stream with emerging new classes under the limited memory environment. No existing methods have the capability to work under the same condition, as far as we know.

  In the future, we plan to improve the proposed method to deal with concept drift and to differentiate two or more emerging new classes before model updates. From a broader perspective, the proposed method is the first implementation of the unsupervised-anomaly-detection-focused approach to the \emph{SENC} problem. We intend to explore other implementations of the same approach.

\section{appendices}
\subsection{Parameter settings}
The parameter settings of all algorithms used in the experiments are provided in Table \ref{parametersettings}. A 10-fold cross-validation on the training set is used in the parameter search to determine the final settings for all SVM algorithms. The parameter search for LOF is as described in \cite{DBLP:conf/aaai/DaYZ14}.
ECSMiner employs $K$-means and $K$ is set to 5 in the experiment.

\begin{table}[h]
  \centering
  \caption{The settings used in the experiments. }
  \label{parametersettings}
  \begin{tabular}{|c|c|}
    \hline
   Method    & Parameter setting \& search range         \\
       \hline
     LOF &  $k= [3,9]$      \\
        \hline
     1SVM & $c=0.1\sim100$, default settings in others   \\
        \hline
     1R-SVM &$\gamma=2^{\alpha} / num\_features $, $\alpha=[-5,5]$ \\ &$c=0.1\sim100$, and default in others  \\
        \hline
  LACU-SVM & $ramp_s = -0.3, \eta = 1.3,$ \\&$\lambda = 0.1, max\_iter = 10$  \\
    \hline
    ECSMiner  & $S = 250, M = 6, K = 5 $ \\ &$q=5, T_{l} = 200$  \\

        \hline
     iForest& $\psi=200, t=100, MinSize= 10$  \\
        \hline
       \emph{SENCForest} &$\psi=200, z=100, \rho = 3,  MinSize=10$\\
    \hline
  \end{tabular}
  \vspace{-2mm}
\end{table}

\subsection{Descriptions of data sets}

{\bf Synthetic}:
We simulate a data stream using a two dimensional synthetic data set as shown below.
It contains 20,000 instances and has four overlapping Gaussian distribution. The first two initial known classes are marked with purple. In the first period, instances of class blue emerge as the first new class. In the second period, instances of class red emerge as the second new class.

   \begin{figure}[h]
  \centering
  \epsfig{file=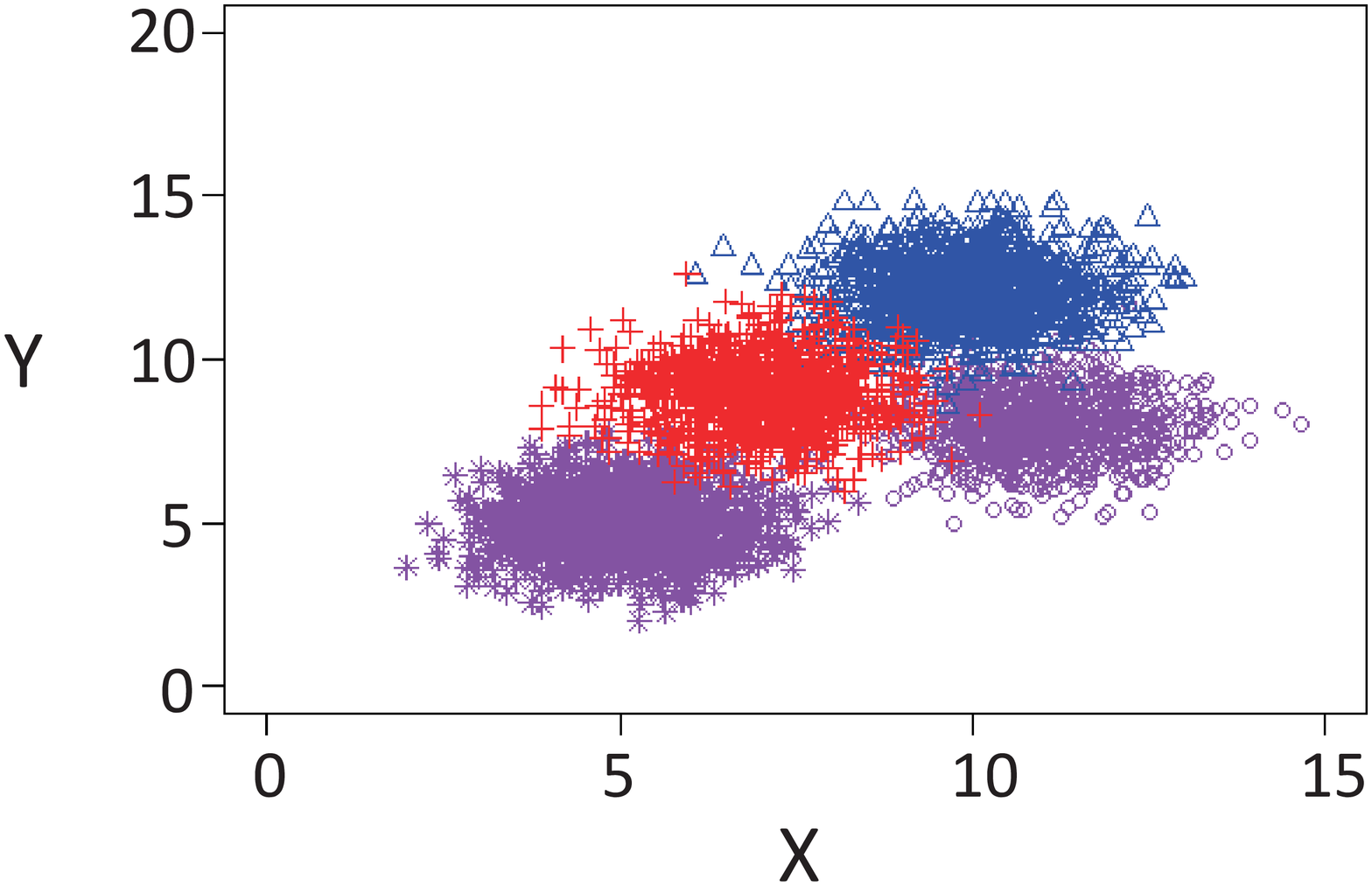,height=1.3in,width=2.2in}
\end{figure}

{\bf MHAR}:
This data set \cite{DBLP:conf/iwaal/AnguitaGOPR12} is collected from 30 volunteers wearing a smart phone on the waist and performing 6 activities (walking, upstairs, downstairs, standing, sitting, laying). The embedded 3D-accelerometer and 3D-gyroscope of a Samsung Galaxy S2 smart phone were used to collect data at a constant rate of 50 Hz. This data set has 6 classes, 10299 instances and 561 attributes.

\bibliography{sigproc}\bibliographystyle{alpha}

\end{document}